%% file: main.tex
\pdfoutput=1

\documentclass[11pt]{article}

\usepackage[final]{acl}

\usepackage{times}
\usepackage{latexsym}

\usepackage[T1]{fontenc}

\usepackage[utf8]{inputenc}

\usepackage{microtype}

\usepackage{inconsolata}

\usepackage{graphicx}

\usepackage{booktabs}
\usepackage{multirow}
\usepackage{tcolorbox}
\usepackage{xcolor}
\usepackage{amsmath}
\usepackage{amsfonts}
\usepackage{graphicx}
\usepackage{subfigure}
\usepackage{subcaption}
\usepackage{caption}
\usepackage{array}

\usepackage{color, xspace}
\usepackage{wrapfig}
\newcommand{\ie}{\emph{i.e.,}\xspace}
\newcommand{\eg}{\emph{e.g.,}\xspace}

\newcommand{\mname}{\textbf{MeCo}\xspace}
\newcommand{\dname}{\textbf{MeCa}\xspace}

\usepackage[normalem]{ulem}
\usepackage{enumitem}
\newtheorem{definition}{Definition}

\title{Adaptive Tool Use in Large Language Models with Meta-Cognition Trigger}

\author{
 \textbf{Wenjun Li\textsuperscript{*,$\dagger$}},
 \textbf{Dexun Li\textsuperscript{*}},
 \textbf{Kuicai Dong\textsuperscript{}},
 \textbf{Cong Zhang\textsuperscript{}},
 \textbf{Hao Zhang\textsuperscript{}},
 \textbf{Weiwen Liu\textsuperscript{}},\\
 \textbf{Yasheng Wang\textsuperscript{}},
 \textbf{Ruiming Tang\textsuperscript{}},
 \textbf{Yong Liu\textsuperscript{}}
\\
 \textsuperscript{}Huawei Noah's Ark Lab
}
\begin{document}
\maketitle
\def\thefootnote{*}\footnotetext{Equal contribution.}
\def\thefootnote{$\dagger$}\footnotetext{Correspondence to wenjunli2017@gmail.com.}

\begin{abstract}
Large language models (LLMs) have shown remarkable emergent capabilities, transforming the execution of functional tasks by leveraging external tools for complex problems that require specialized processing or up-to-date data. While existing research expands LLMs access to diverse tools (\eg program interpreters, search engines, calculators), the necessity of using these tools is often overlooked, leading to indiscriminate tool invocation. This naive approach raises two key issues: increased latency due to unnecessary tool calls, and potential errors resulting from faulty interactions with external tools. In this paper, we introduce meta-cognition as a proxy for LLMs self-assessment of their capabilities, reflecting the model's awareness of its own limitations. Based on this, we propose {\em MeCo}, an adaptive decision-making strategy for external tool use. MeCo quantifies metacognitive scores by capturing high-level cognitive signals in the representation space, guiding when to invoke tools. Notably, MeCo is fine-tuning-free and incurs minimal cost. Experiments across multiple backbone models and benchmarks show that MeCo reliably detects LLMs' internal cognitive signals and significantly improves tool-use decision-making.
\end{abstract}

\section{Introduction}
Equipping large language models (LLMs) with tool use capabilities allows them to overcome their limitations by accessing external or up-to-date information~\citep{komeili2021internet,tang2023toolalpaca}, domain-specific knowledge~\citep{he2023solving,schick2024toolformer}, and advanced specialized functionalities~\citep{yang2023foundation,gao2023pal,lu2024chameleon}, thereby enabling them to handle more complex tasks beyond their inherent abilities. While prior research has focused on expanding the tool arrays~\citep{qin2023toolllm,hao2024toolkengpt} and optimizing their use~\citep{patil2023gorilla,shen2024hugginggpt}, the decision-making process for determining when tools are necessary remains underexplored.

Naively invoking tools indiscriminately leads to two major issues: (1) increased latency~\citep{qu2024tool,wang2024survey}, as external tools, such as search engine, typically operates slower compared to relying on internal knowledge retrieval, and (2) robustness risks, where dependence on external APIs increases the likelihood of errors due to tools malfunction or unnecessary tool use~\citep{qin2023toolllm,lu2024chameleon,wurepoformer}.

\begin{figure*}[th]
  \centering
  \includegraphics[width=\linewidth]{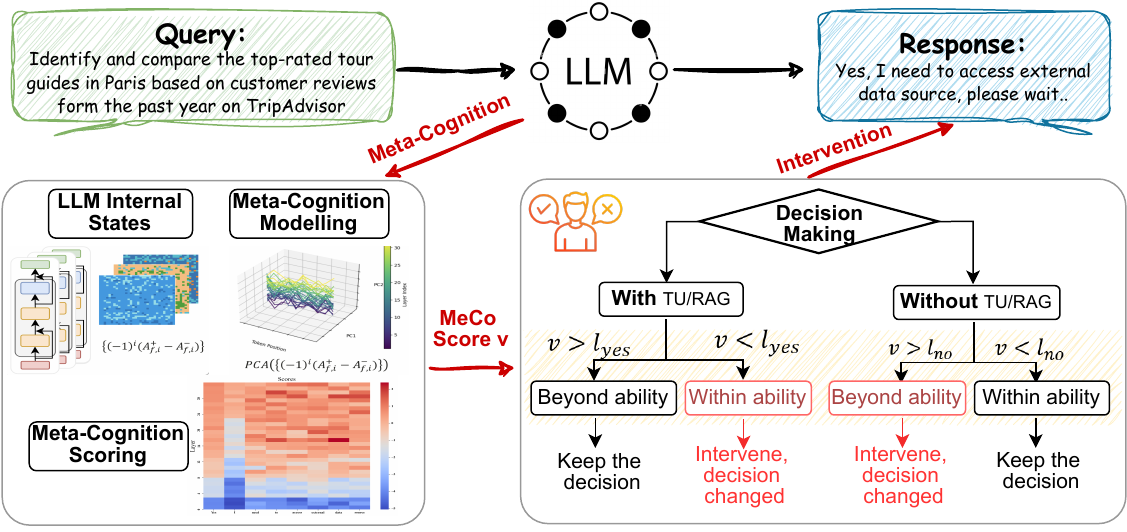}
  \caption{Overview of \mname: Learned Meta-Cognition determines the necessity for tool use or retrieval by using a trained meta-cognition probe to detect the internal state of an LLM.}
  \label{fig:algo_overview}
\end{figure*}

To address these limitations, we propose an adaptive tool-use strategy that improves decision-making in LLMs. Our approach, \mname (\textbf{Me}ta-\textbf{Co}gnition-oriented trigger), enables LLMs to self-assess their capabilities and decide whether external tools are needed to answer a given query. \mname incorporates three key principles:
\begin{itemize}[leftmargin=*, itemsep=0em, topsep=-0.0em]
    \item \textbf{Meta-Cognition:} 
    We define {\em meta-cognition} in the context of LLM tool useas the model’s ability to evaluate its own competence based on internal representations. This self-awareness is crucial for minimizing unnecessary tool use.
    \item \textbf{Effective Strategy Utilization:} We design an efficient strategy that leverages the quantified meta-cognition signals to dynamically adjust tool use decisions, significantly improving decision accuracy compared to baseline methods.
    \item \textbf{Generability:} We demonstrate that \mname\ generalizes well across diverse tasks and domains. Additionally, we treat adaptive Retrieval-Augmented Generation (RAG) as a special case of tool use and show \mname outperforms standard baselines.
\end{itemize} 

Building on the Representation Engineering (RepE)~\citep{zou2023representation}, we develop a computationally efficient plug-in module to assess meta-cognition in LLMs. Our analysis reveals that meta-cognition provides strong and interpretable signals that can effectively guide tool use decisions. As illustrated in Figure~\ref{fig:algo_overview}, \mname employs a dual-thresholding policy to distinguish between strong and weak meta-cognitive signals, refining decision-making when uncertainty arises.

In summary, our contributions are four-fold: 1) We introduce the concept of adaptive tool use, which enhances both the efficiency and robustness of LLM tool use paradigms. 2) We unify adaptive tool use and adaptive RAG under a shared framework, with their activation driven by meta-cognition-based decision-making. 3) We build a new benchmark, \dname, to systematially evaluate the effectiveness of our approach. 4) We empirically demonstrate that \mname significantly enhances model awareness in both tool utilization and RAG scenarios.

\begin{figure*}[th]
  \centering
  \includegraphics[width=1\linewidth]{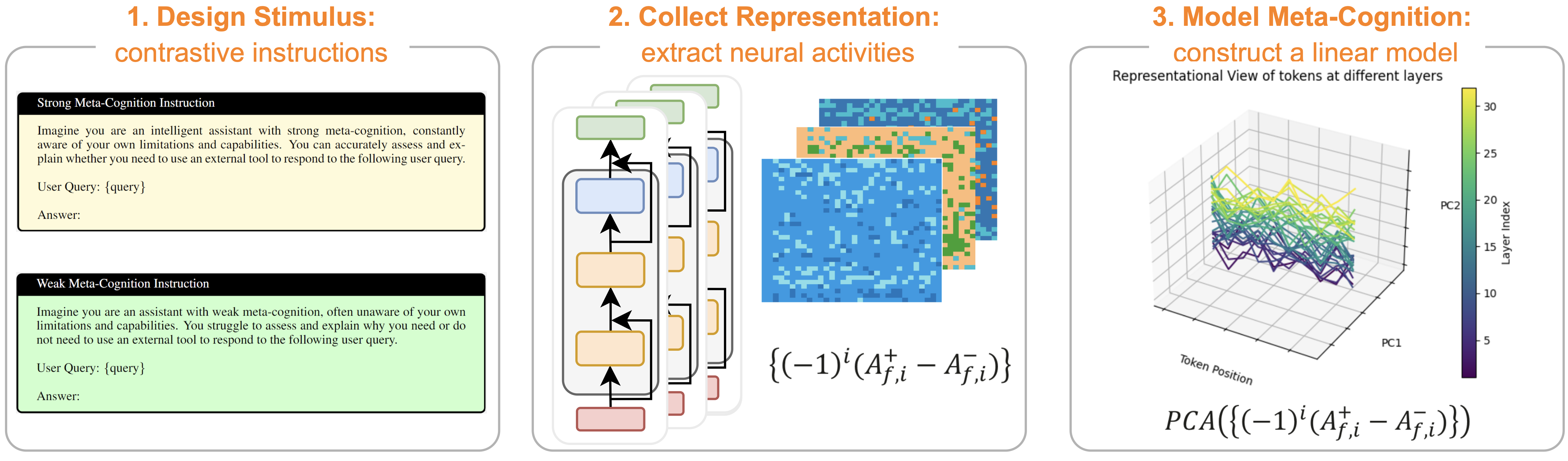}
  \caption{Pipeline for training the meta-cognition probe.}
  \label{fig:pipeline_probe_training}
\end{figure*}

\section{Background} \label{sec:background}
Recent studies have explored the internal representations of LLMs to improve interpretability and understand their implicit beliefs~\citep{bricken2023monosemanticity,Levinstein_2024}. Prior work~\citep{zou2023representation, liu2023aligning} shows that abstract features like happiness, honesty, and confidence correspond to distinct, linearly separable directions in the representation space. Figure \ref{fig:pipeline_probe_training} illustrates the pipeline for training these feature probes. To detect these signals, contrastive instruction pairs are typically used to induce their emergence.

Building on these findings, we can capture and manipulate high-level functions $f$ (\eg honesty) in model responses. We follow \citet{zou2023representation} to design an \textit{experimental prompt} $T^{+}_{f}$ that requires function execution and a \textit{reference prompt} $T^{-}_{f}$ that does not. The instruction template is as follows:

\begin{tcolorbox}[colback=white!95!white, colframe=black, width=\columnwidth, arc=0mm, outer arc=0mm]
\small
\textbf{USER:} \textcolor{blue}{$\langle$ instruction $\rangle$} \textcolor{red}{$\langle$ experimental / reference prompt $\rangle$} \\
\textbf{ASSISTANT:} \textcolor{blue}{$\langle$ output $\rangle$}
\end{tcolorbox}

For a function $f$ and model $M$, given instruction-response pairs $(q_i, a_i)$ in set $S$ and denoting a response truncated after token $k$ as $a^k_i$, we collect internal representations for the experimental and reference sets:
\begin{equation}
    A^{\pm}_f = \left\{\text{Rep}(M, T^{\pm}_{f}(q_i, a^k_i))[-1] \text{ }|\text{ } (q_i, a_i) \in S \right\} \label{eq:RepE}
\end{equation}
where Rep denotes the representation obtaining operation, $[-1]$ is the last token representation of $a^k$, and $A^{\pm}_f$ are the resulted activations consist of individual vectors. 

The goal is to learn a linear model to predict the direction of the function $A^{\pm}_f$ based on internal representations. Specifically, we apply PCA~\citep{mackiewicz1993principal} in an unsupervised manner to pairwise difference vectors, deriving the first principal component $v_f$ (referred to as the {\em probe}) to identify function directions in the representation space. Equation (\ref{eq:RepE}) is applied at each layer of $M$ to derive layer-wise probes which are then used to interact with the LLM's representations to monitor and control its behavior.

\section{Approach}
We define meta-cognition in LLMs as follows:
\begin{definition} 
Meta-cognition refers to an LLM's ability to assess and regulate its own knowledge and limitations, enabling informed decision-making about task execution, including when to rely on external tools or resources.
\end{definition}
In the context of tool use, this involves assessing the model’s capabilities and limitations to determine whether a query can be answered independently or requires external tools, based on the complexity of the query and the sufficiency of model’s internal knowledge.

To quantify meta-cognition, we train a probe that detects the model's level of meta-cognitive awareness. This probe evaluates the rationale behind the model's decision-making process, providing a score that reflects the model's self-assessment accuracy.
For instance, when the model receives a complex mathematical query, the meta-cognition probe assesses whether it correctly decides to solve the problem itself or delegate it to a calculator. A high meta-cognition score indicates accurate self-assessment, while a low score suggests uncertainty in the model's decision.

\subsection{Meta-Cognition Probe Extraction}
Training a meta-cognition probe differs fundamentally from training honesty or confidence probes. The latter probes are typically trained on true/false factual statements, \eg "fire needs oxygen to burn" and "oxygen is harmful to human breathing". These statements are independent of specific user queries, meaning the model produces consistent internal representations regardless of the prompt context.

In contrast, detecting internal cognitive signals related to tool use requires query-dependent responses. To achieve this, we use leading proprietary LLMs to generate user queries related to tool use and their corresponding responses (\ie binary Yes/No answers with brief explanations). We then construct the training dataset following the procedures outlined in Section \ref{sec:background}. 

Notably, only a small dataset suffices for strong probe performance. Appendix~\ref{app:sec:probe} provides an analysis of the relationship between probe performance and training data size. Specifically, after collecting the instruction-response pairs $(q_i,a_i)$, we extract the model's internal representations and compute the contrastive representation differences $A^{\pm}_f$ according to Equation (\ref{eq:RepE}). We then apply PCA to the set of difference vectors $\{(-1)^i(A_{f,i}^+-A_{f,i}^-)\}$ to obtain the first principal component $\nu_f$ as the meta-cognition probe, which identifies the direction of the underlying meta-cognition signal.

\begin{figure}[ht]
  \centering
  \includegraphics[width=1.0\linewidth]{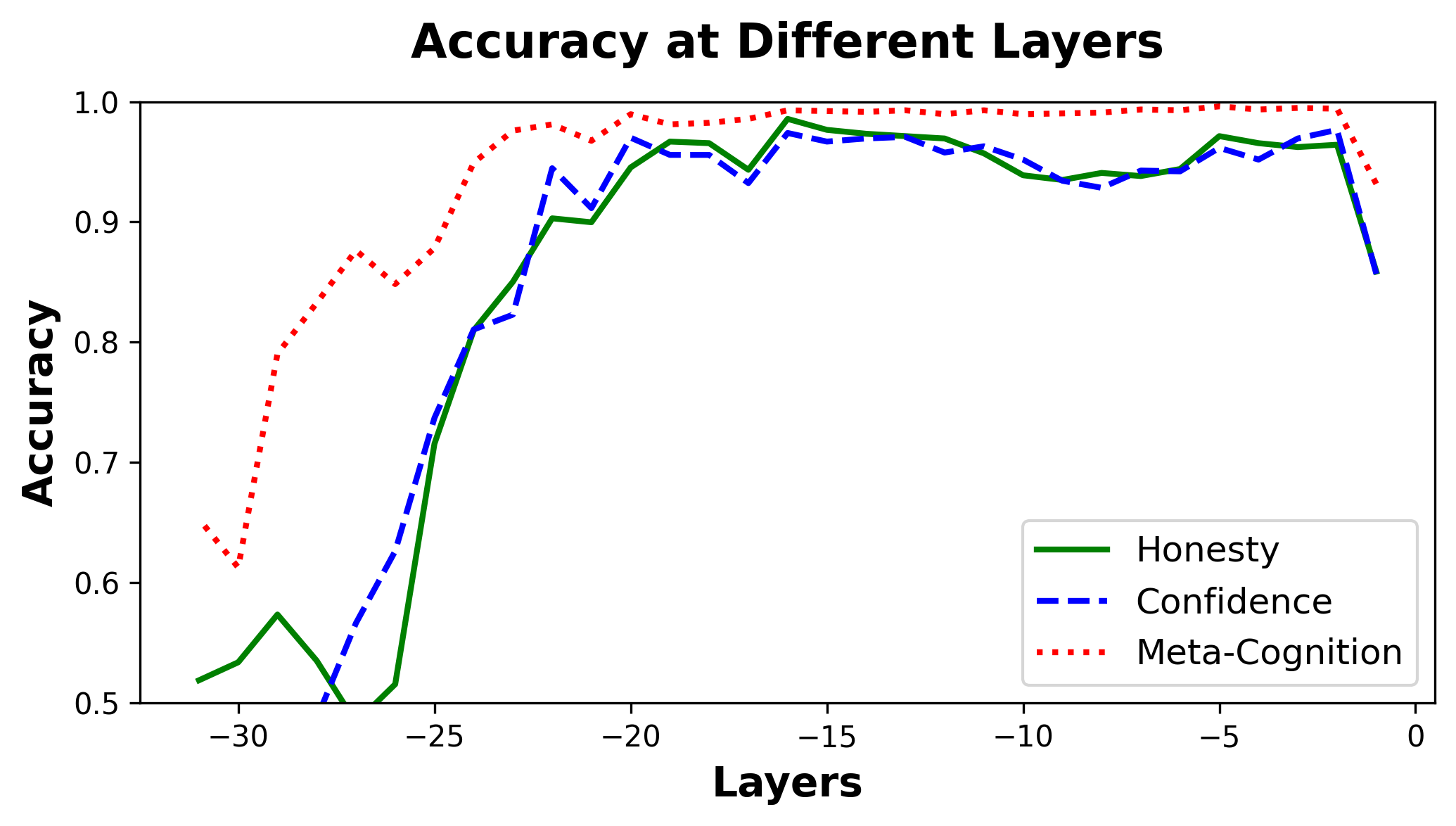}
  \caption{Comparison between different probes. Note that -1 means the last layer in the LLMs.}
  \label{fig:probes_comparison}
\end{figure}

After the above procedures, we obtain meta-cognition probes across all model layers (\eg 32 probes for \texttt{Llama-3-8B}). We evaluate our meta-cognition probe against existing probes for honesty \cite{zou2023representation} and confidence \cite{liu2024ctrla} by measuring the {\em intermediate classification accuracy} on held-out examples where the model is instructed to exhibit either honest/confident/strong meta-cognition or dishonest/lacking confidence/weak meta-cognition. Intermediate classification accuracy refers to the probe’s ability to correctly predict the directional alignment, \ie whether the token representation points in a positive or negative direction along the concept axis (honesty/confidence/meta-cognition). This metric does not reflect final tool-use decision accuracy. The results shown in  Figure~\ref{fig:probes_comparison} indicates our meta-cognition probe achieves near-optimal accuracy, significantly outperforming prior probes. 

\subsection{Decision-Making Strategy based on Meta-Cognition}
With an accurate meta-cognition probe, we design a decision-making strategy based on the detection results. For a given query, the LLM generates a response of $m$ tokens, each with a meta-cognition score across $n$ layers, forming a detection array of size $(m, n)$. The final decision—"Yes" (use external tools/RAG) or "No" (respond independently)—is derived from this array.

\paragraph{Reducing $m$ to 1.}
We examine various prompting strategies (detailed in Appendix~\ref{app:sec:extend_results}) and find that the Yes/No+Explanation strategy, where the model answers with "Yes/No" followed by a brief explanation, yields the best performance. Hence, we focus on the first token of the model's response, as it provides a clear signal of whether the model decides to rely on external tools. Extracting the meta-cognition score of the first token to represent the whole response simplifies our decision-making process, since calculating an overall meta-cognition score for the entire response is challenging due to varying response lengths and content across different queries. As the model consistently responds with "Yes/No" as the first token, basing the trigger mechanism on the first token's meta-cognition score is both reasonable and practical.

\paragraph{Reducing $n$ to 1.}
In \citet{zou2023representation} and \citet{liu2024ctrla}, a mean score from multiple probes' results is typically used to represent the token's final quantification. However, our experiments show that scores predicted by different probes vary significantly, and simply averaging them does not yield accurate results. We found that probes in shallower layers (\eg layers -5 to -2) tend to be more effective, with appropriate score distributions, ranges, and lower variances. Therefore, we select the single probe with the highest classification accuracy within layers -5 to -2 (as shown in Figure~\ref{fig:probes_comparison}) as the final predictor.

\begin{figure*}[th]
    \centering
    \subfigure[Correct Yes/No]{
        \includegraphics[width=0.31\textwidth]{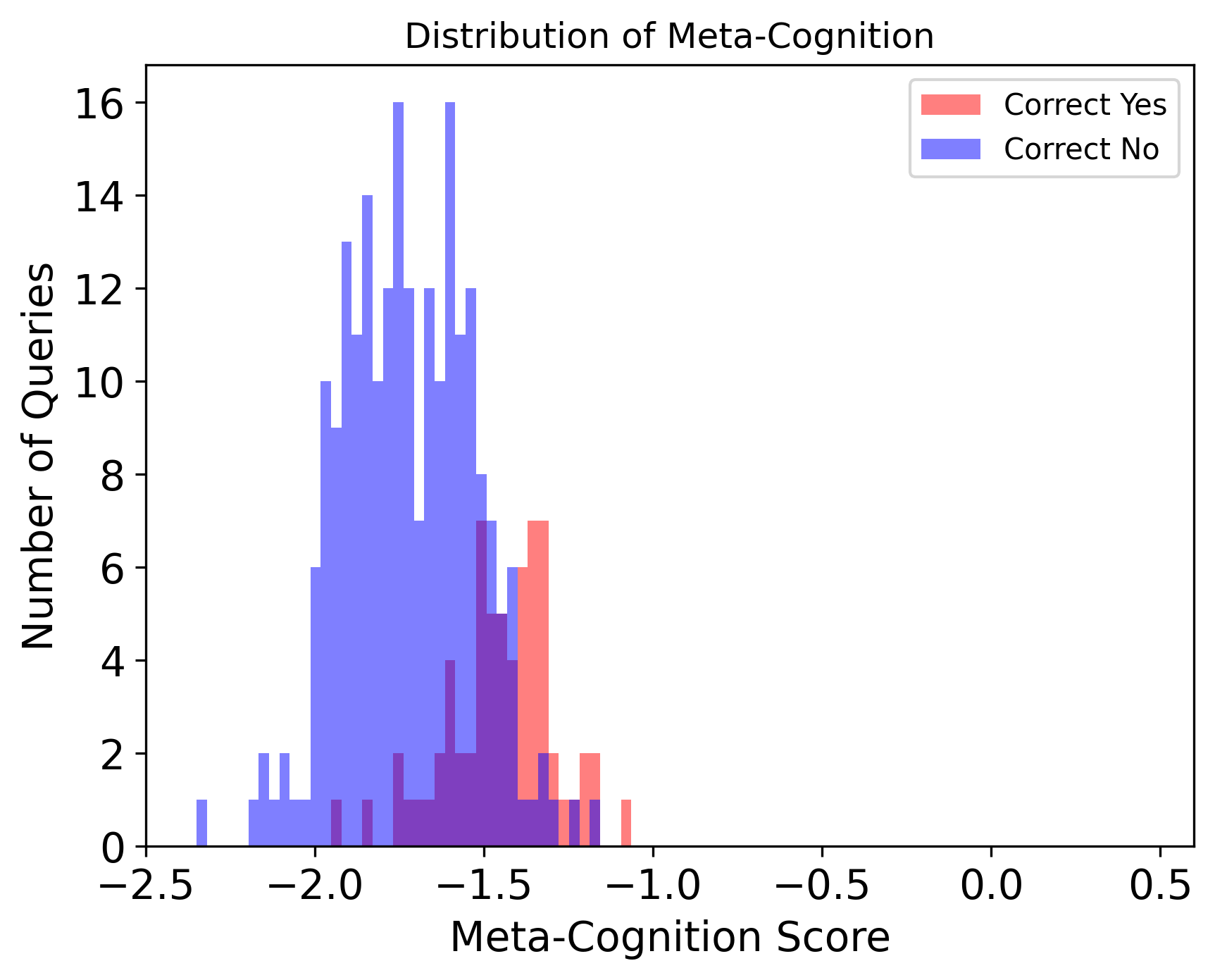}
    }
    \hfill
    \subfigure[Correct/Incorrect Yes]{
        \includegraphics[width=0.31\textwidth]{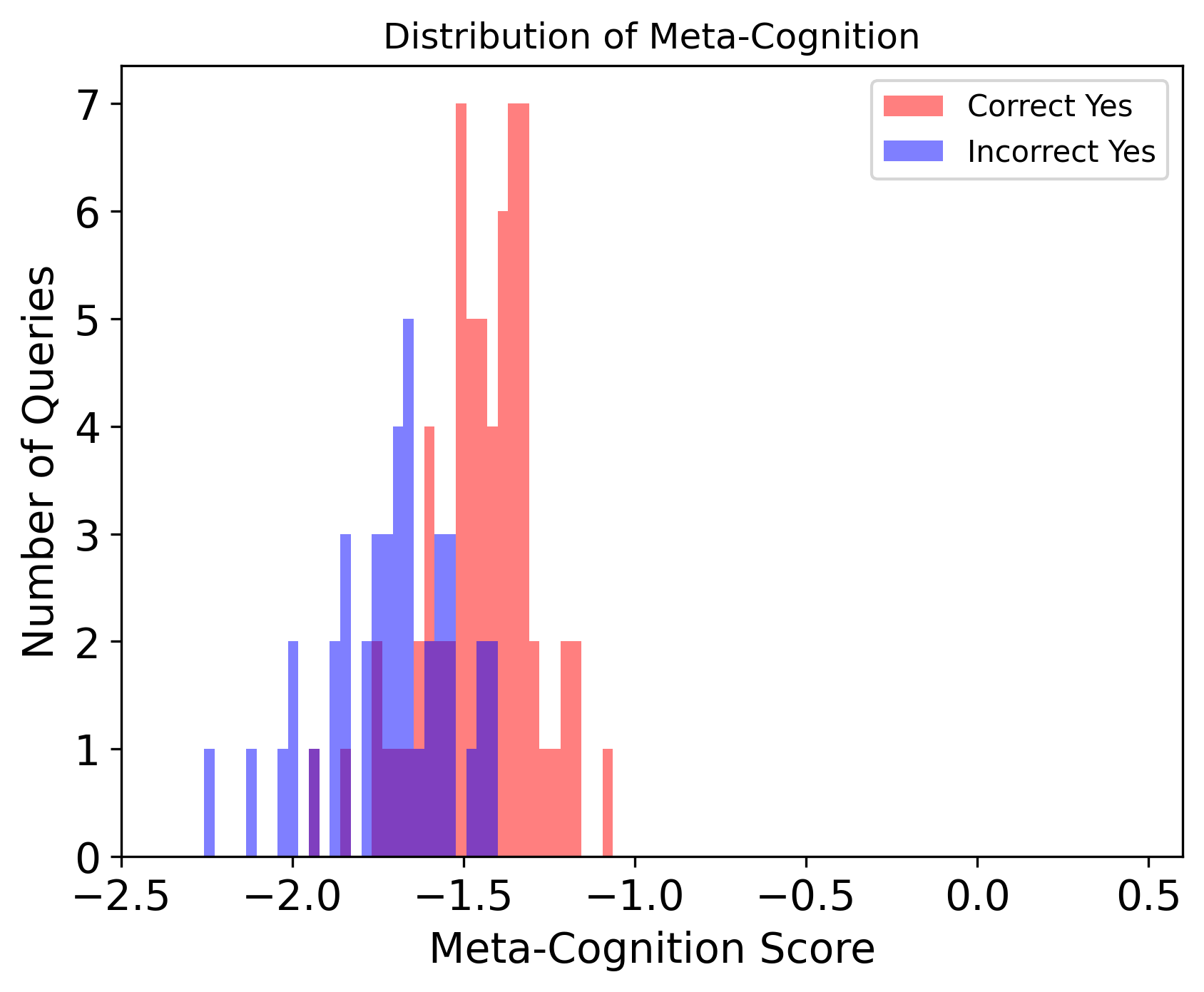}
    }
    \hfill
    \subfigure[Correct/Incorrect No]{
        \includegraphics[width=0.31\textwidth]{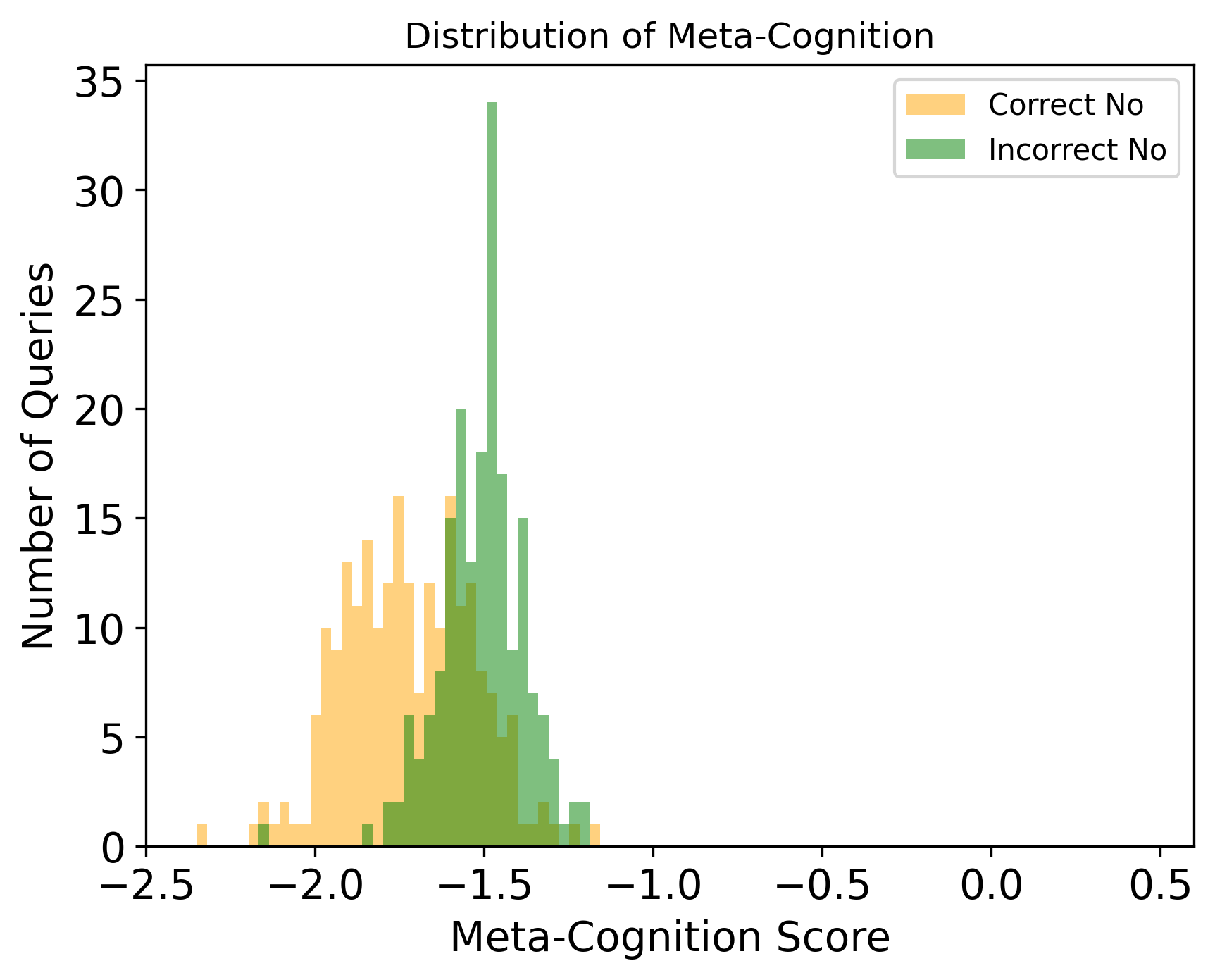}
    }
    \vskip\baselineskip
    \subfigure[Correct Yes/No]{
        \includegraphics[width=0.31\textwidth]{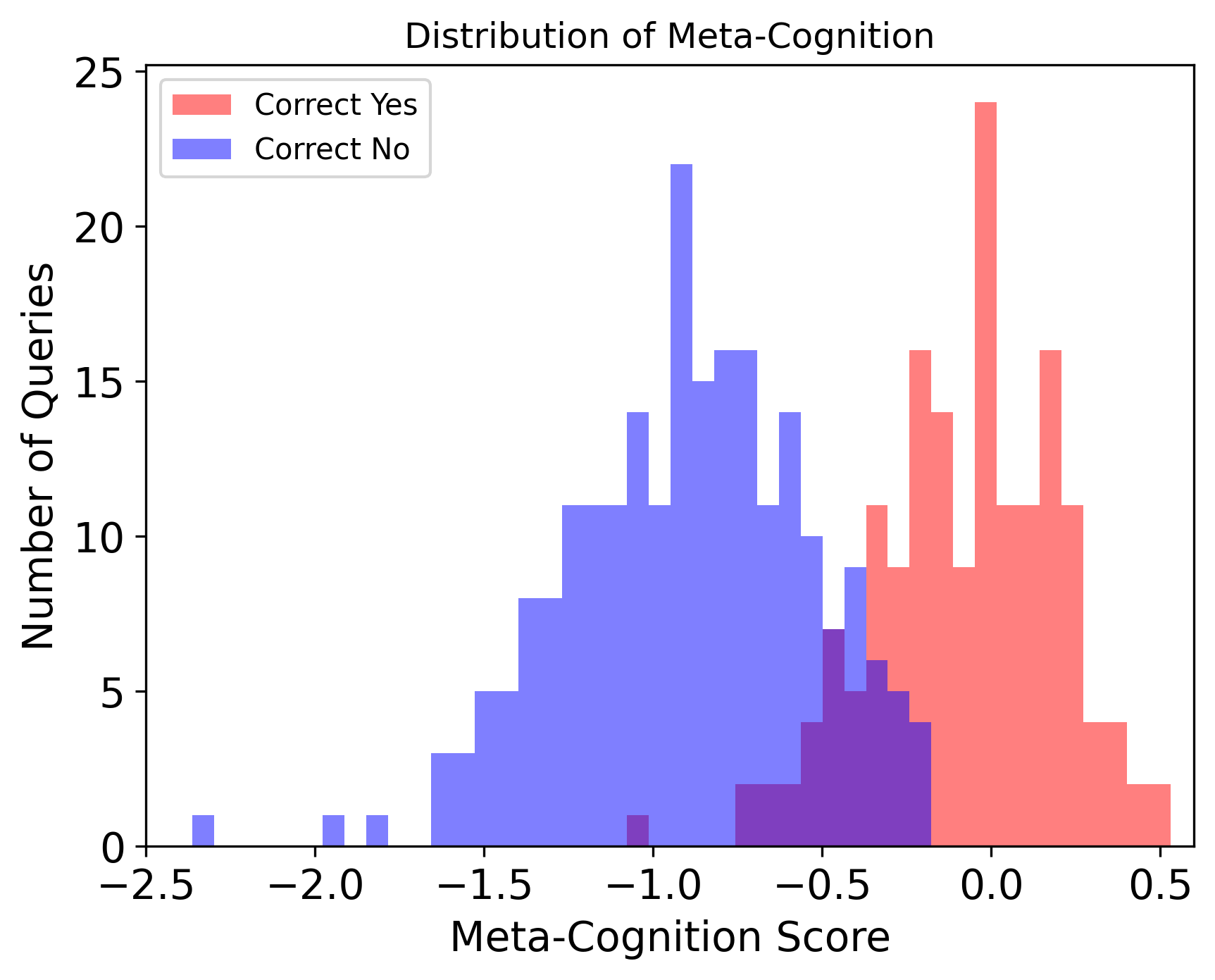}
    }
    \hfill
    \subfigure[Correct/Incorrect Yes]{
        \includegraphics[width=0.31\textwidth]{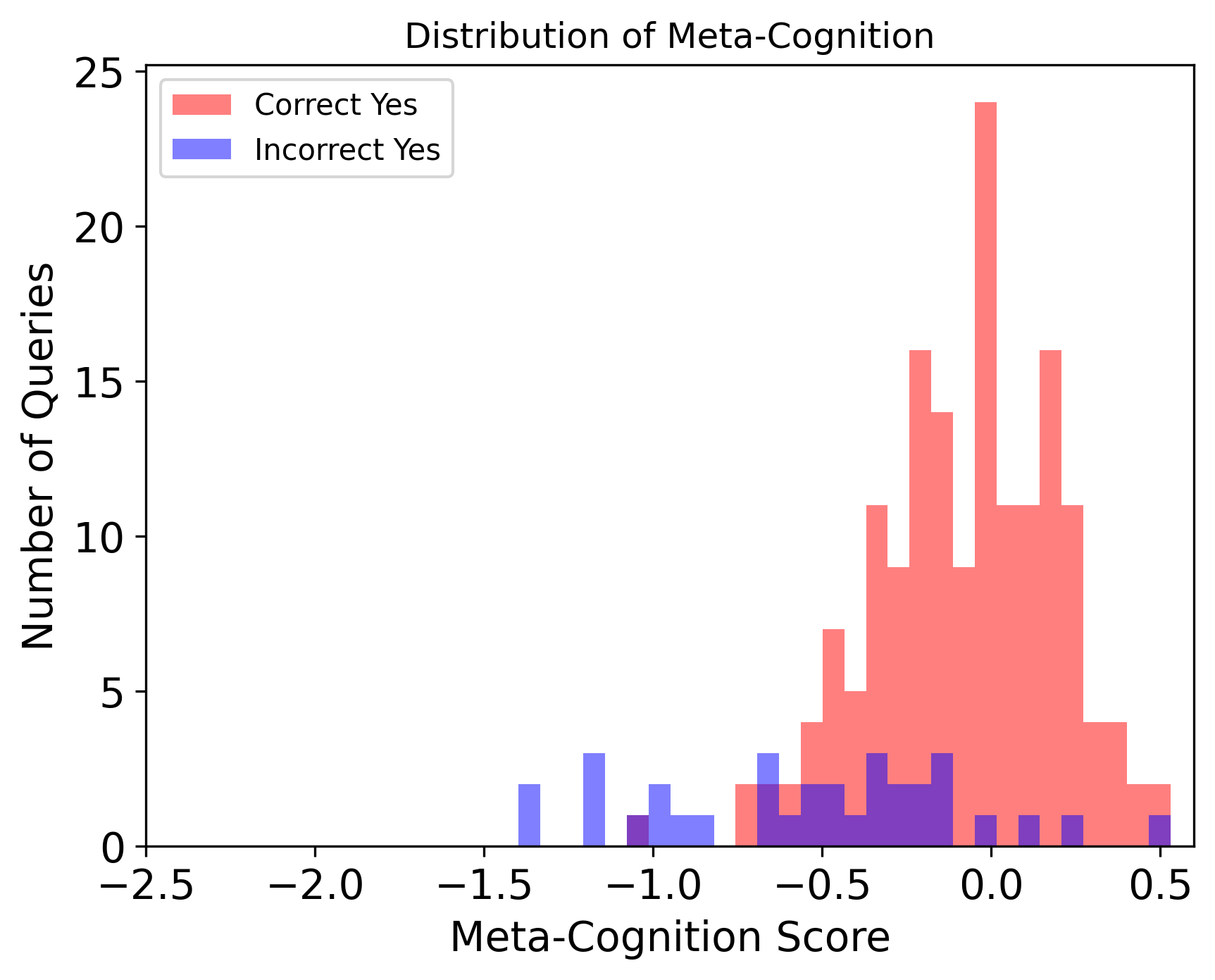}
    }
    \hfill
    \subfigure[Correct/Incorrect No]{
        \includegraphics[width=0.31\textwidth]{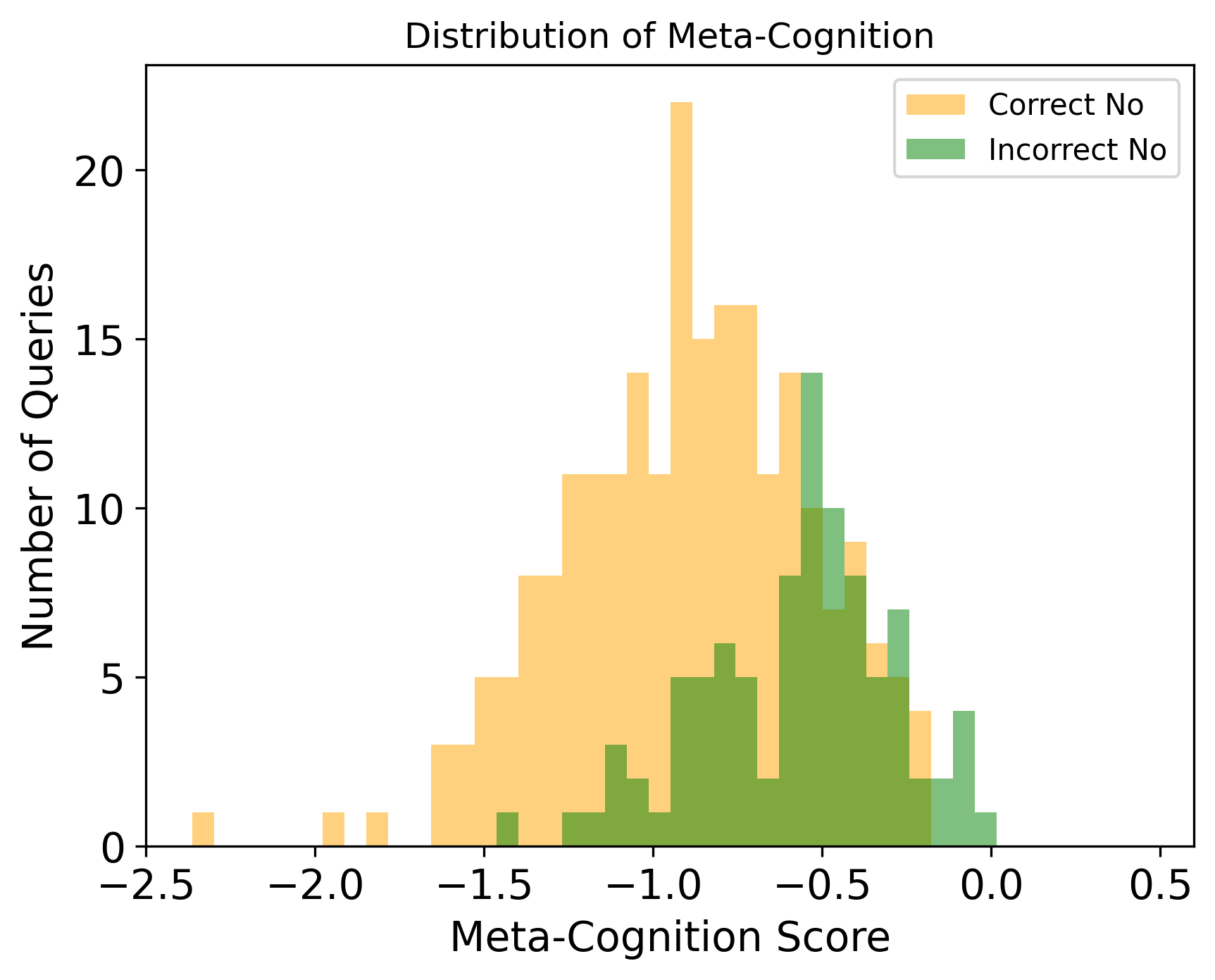}
    }
    \caption{Distribution of meta-cognition scores of the first token in model responses. (a), (b), and (c) are from \texttt{LM3-8B} (pre-fine-tuning), while (d), (e), and (f) are from \texttt{LM3-8B-sft} (post-fine-tuning). The scores are derived from the train data in MetaTool, using prompts without context.}
    \label{fig:metacog_distribution}
\end{figure*}

\paragraph{Dual-Thresholding.}
After distilling the meta-cognition scores into a single scalar value, we adopt a simple yet effective dual-thresholding strategy (as illustrated in Figure~\ref{fig:algo_overview}) to determine the optimal thresholds, $l_{yes}$ and $l_{no}$, using validation data. These thresholds are then directly applied to the test data. As shown in Figure~\ref{fig:metacog_distribution}, we empirically observe that correct Yes responses tend to have higher meta-cognition scores than incorrect Yes responses, while correct No responses tend to have lower scores than incorrect No responses. Notably, in both pre- and post-fine-tuning experiments, there is a clear gap between the meta-cognition scores of correct and incorrect responses. Our decision-making strategy identifies and leverages this gap to make better decisions. This pattern arises because the meta-cognition score is influenced by the token embedding of Yes/No, making scores between the two classes not directly comparable. Consequently, Yes scores should only be compared with other Yes scores, and similarly for No scores.

Driven by this empirical observation, we retain the model’s decision when the meta-cognition score $\nu$ is higher than $l_{yes}$ for Yes decisions, or lower than $l_{no}$ for No decisions; and override the model’s decision when $\nu$ is lower than $l_{yes}$ for Yes decisions, or higher than $l_{no}$ for No decisions. We demonstrate the robustness and effectiveness of this dual-thresholding strategy through comprehensive evaluations on both the MetaTool and \dname benchmarks in Section~\ref{sec:experiments}.

\begin{figure*}[t]
  \centering
  \includegraphics[width=1\linewidth]{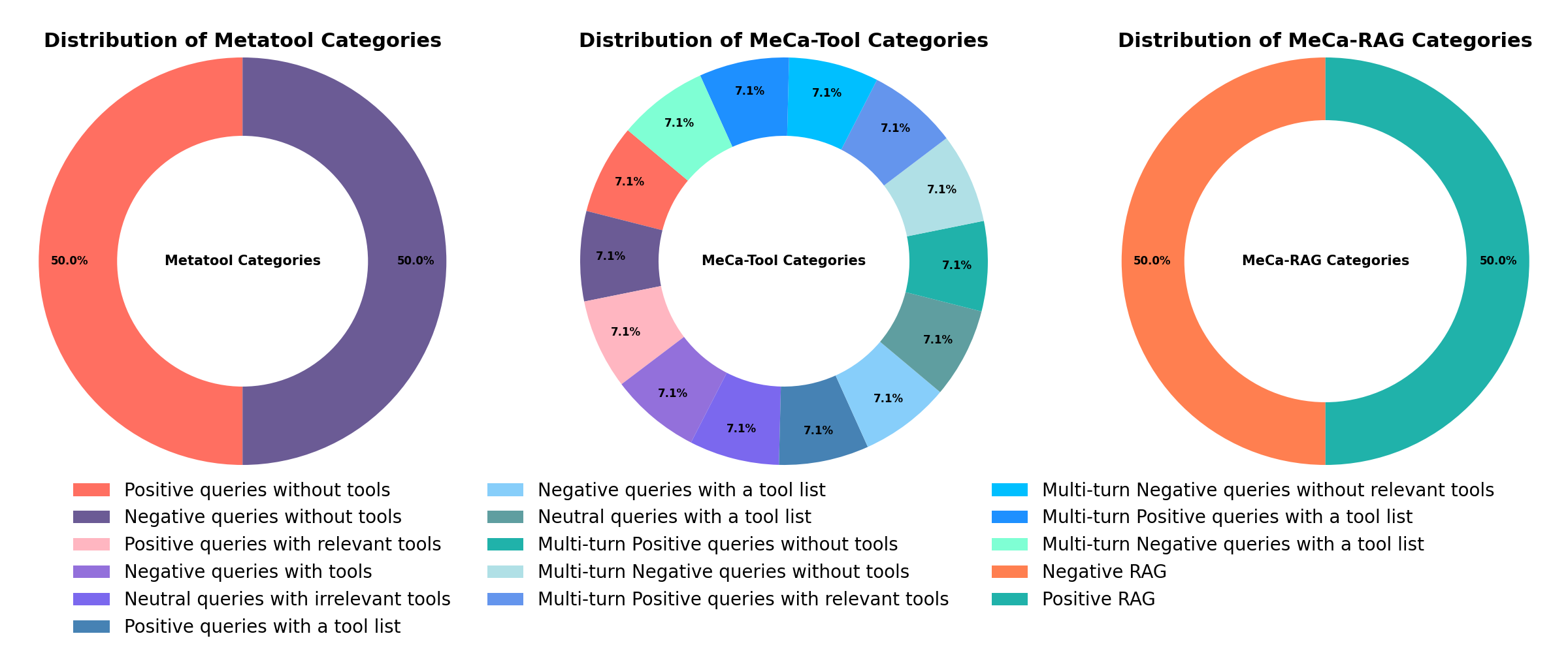}
  \caption{Overview of benchmarks: Distribution of MetaTool, \dname-Tool, and \dname-RAG categories. MetaTool and \dname-Tool assess the necessity of tool use, while \dname-RAG evaluates the necessity of RAG interactions.}
  \label{fig:MeCa}
\end{figure*}

\section{Benchmark-\dname}
We evaluate \mname using MetaTool~\citep{huang2023MetaTool} and introduce a new benchmark, Meta-Cognitive Tool Assessment (\dname), in which each query undergoes human review. \dname extends MetaTool by incorporating a broader range of scenarios to assess adaptive tool use and RAG.

MetaTool comprises 1,040 queries designed to evaluate whether LLMs recognize when to use external tools. In MetaTool, LLMs must decide on tool usage based solely on user queries, without tool names or descriptions. We identify the following limitations of MetaTool: 1) queries lack supplementary information or explicit tool provisions, whereas real-world tasks involve more complex intents and diverse requirements; 2) it primarily focuses on single-turn tool usage decisions. 

To address these gaps and provide a more robust evaluation of \mname, we introduce \dname, which includes two main components: \dname-Tool and \dname-RAG. \dname-Tool extends MetaTool by expanding tool-related assessments across three key categories:
\begin{itemize}[leftmargin=*] 
\item \textbf{Tool Use Assessment:} Evaluates whether the LLM should invoke external tools.
    \begin{itemize} 
    \item Queries solvable by the LLM without tools.
    \item Queries requiring one or more tools due to insufficient internal capabilities.
    \end{itemize}
\item \textbf{Provided Tool Evaluation:} Tests the LLM’s ability to decide on tool usage when given a predefined set of tools.
    \begin{itemize} 
    \item Cases where external tools are unnecessary.
    \item Cases where essential tools are available and should be used.
    \item Cases where required tools are missing.
    \end{itemize}
\item \textbf{Multi-turn Interaction:} Assesses tool use decisions in multi-turn dialogues that require adaptation to evolving contexts.
    \begin{itemize}
    \item Cases where external tools are unnecessary.
    \item Cases where essential tools are available and should be used.
    \item Cases where required tools are missing.
    \end{itemize}
\end{itemize}

Specifically, we create six evaluation tasks in \dname-Tool to systematically assess an LLM’s ability to make tool-related decisions across various scenarios. Tasks 1 and 4 evaluate whether an external tool is necessary to solve a given query. Tasks 2 and 5 assess the LLM’s ability to determine the relevance of a provided tool, including cases where the tool is irrelevant. Tasks 3 and 6 further extend this evaluation by presenting multiple tools (ranging from 2 to 5) and requiring the LLM to select the appropriate one. Notably, Tasks 1–3 focus on single-turn settings, while Tasks 4–6 involve multi-turn dialogues, testing the LLM's ability to adapt tool-use decisions to evolving conversational contexts. \dname-Tool significantly expands MetaTool by covering six tasks with 7,000 queries, providing a more diverse and comprehensive evaluation framework. The query composition of MetaTool and \dname-Tool is illustrated in Figure~\ref{fig:MeCa}, and detailed task statistics are provided in Table~\ref{tab:MeCa} in Appendix~\ref{app:sec:meca}.

Beyond tool use, \dname-RAG evaluates adaptive RAG—determining whether a query can be answered directly by the LLM or requires external retrieval. RAG is a special case of tool use, where the LLM’s internal knowledge is insufficient and necessitates using of a search engine to access external information. \dname-RAG includes:
\begin{itemize}[leftmargin=*] 
\item \textbf{Positive RAG:} Cases where retrieval is essential to answer complex queries or those requiring up-to-date information.
\item \textbf{Negative RAG:} Cases where the LLM can provide correct answers using its internal knowledge without retrieval.
\end{itemize}

To curate the \dname-Tool dataset, we employed a meticulous approach that began with collecting diverse scenarios from various online corpora, ensuring our synthetic APIs and conversations were grounded in realistic contexts. Based on these scenarios, we designed 500 distinct synthetic APIs emulating real-world applications across various domains. We then generated queries by randomly sampling from this API pool—queries that either require tool invocation, rely on the LLM’s internal knowledge, or expose cases where the provided APIs are insufficient—followed by rigorous human verification for accuracy. For the \dname-RAG dataset, we selected common fact-based data from the RepE~\citep{zou2023representation} dataset (\eg "The Earth orbits the Sun”) to generate negative queries that do not require retrieval, while positive queries were generated from crawled recent news unseen by LLMs, thereby necessitating retrieval of new information. 

By incorporating both adaptive tool usage and adaptive RAG, \dname serves as a robust benchmark for assessing LLM decision-making in complex scenarios. Detailed benchmark statistics and creation methodology for \dname-RAG can be found in Appendix~\ref{app:sec:meca}.

\section{Experiment Setup}\label{sec:setup}
\textbf{Baselines}: We evaluate \mname against two baselines: Naive and $P_{\text{Yes}}$. The Naive baseline determines "Yes" or "No" based solely on the first token generated by the LLM, where "Yes" indicates the need for external tools, and "No" indicates otherwise. The $P_{\text{Yes}}$ baseline refines this by computing a {\em Yes-score}, 
$$
\text{Yes-score} = \frac{P(\text{Yes }|\text{ Prompt})}{P(\text{Yes }|\text{ Prompt}) + P(\text{No }|\text{ Prompt})} 
$$
which ranges from 0 (full "No") to 1 (full "Yes"), with values near 0.5 indicating uncertainty. Instead of relying solely on the first token, $P_{\text{Yes}}$ learns an optimal threshold: scores above this threshold are classified as "Yes," while those below are classified as "No." Further details are in Section~\ref{app:subsec:p_yes}.

\textbf{Backbone LLMs}: We employ \textit{Llama-3-8B-Instruct}, \textit{Mistral-7B-Instruct-v0.3}, and \textit{Llama-3-70B-Instruct} as backbone models to evaluate \mname. For brevity, we refer to them as \texttt{LM3-8B}, \texttt{Mist-7B}, and \texttt{LM3-70B}. Additionally, we fine-tune these models on a dataset of 4,000 tool-use query-response pairs generated by GPT-4-turbo, denoting the fine-tuned versions as \texttt{LM3-8B-sft}, \texttt{Mist-7B-sft}, and \texttt{LM3-70B-sft}. We include the 70B variant to demonstrate \mname’s effectiveness on large-scale models and its potential applicability in industry-grade deployments.

\textbf{Evaluation}: The primary evaluation metric is decision accuracy—whether the model correctly identifies when external tools/RAG are genuinely necessary. Additional metrics (precision, recall, etc.) are reported in Appendix~\ref{app:sec:extend_results}.

\textbf{Prompting Strategies}: We experimented with various prompting strategies, including "Yes/No" with and without explanations and Chain-of-Thought (CoT; \cite{wei2022chain}). The best performance was achieved using the "Yes/No + Explanation" strategy, which is used throughout this paper. Full results for all prompting variants are provided in Appendix~\ref{app:sec:extend_results}. We further employ two types of prompts in our experiments: 1) prompts with context, which include specific reasons why LLMs require external tools to complete user tasks, plus five randomly sampled in-context examples to guide the model's decisions; and 2) prompts without context, a concise version containing only the instruction and query. Exact prompt templates are detailed in Appendix~\ref{app:sec:probe}.

\section{Experiments} \label{sec:experiments}
We conduct extensive experiments to empirically reveal the effectiveness of \mname on MetaTool and \dname. Specifically, we assess \mname in adaptive tool use on both MetaTool and \dname-Tool and in adaptive RAG on \dname-RAG. 

\subsection{Reduced Tool Invocations and Latency}
Before delving into detailed benchmark performance, we first demonstrate that using \mname in adaptive tool use and adaptive RAG reduces unnecessary tool invocations and overall latency. For this evaluation, we use the \texttt{LM3-8B} model.

We begin by evaluating the number of tool invocations on MetaTool, which contains an equal proportion of queries that require and do not require external tools. Table~\ref{tab:num_tool_invocations} summarizes the number of tool invocations made by each method. The Naive method invokes 296 tool calls before fine-tuning and 277 after fine-tuning. $P_{\text{Yes}}$ improves efficiency by reducing the number of invocations to 208 and 273, respectively. Notably, \mname achieves the lowest number of tool invocations—206 before fine-tuning and 240 after—while maintaining the highest decision accuracy. These results highlight \mname's effectiveness in addressing the two key issues associated with indiscriminate tool use:
\begin{itemize}[leftmargin=*]
    \item \textbf{Reduced Latency:} By minimizing unnecessary tool calls, \mname lowers response latency, particularly for queries that do not require external tools (\eg retrieval tools can take up to 14.8 seconds, as shown in Table~\ref{tab:increased_latency}). 
    \item \textbf{Improved Robustness:} Avoiding unnecessary tool invocations reduces reliance on potentially unreliable external APIs, thereby mitigating risks and malfunctions caused by unnecessary tool use.
\end{itemize}

\begin{table}[t]
\small
\centering
\begin{tabular}{l|c|c}
    \toprule
    \textbf{Method} & \textbf{Pre Fine-tuning} & \textbf{Post Fine-tuning} \\
    \midrule
    Always Call           & 520 (50.0\%) & 520 (50.0\%) \\
    Naive                 & 296 (61.9\%) & 277 (82.1\%) \\
    $P_{\text{Yes}}$      & 208 (63.5\%) & 273 (81.7\%) \\
    \mname                & \textbf{206 (65.0\%)} & \textbf{240 (84.3\%)} \\
    \bottomrule
\end{tabular}
\caption{Number of tool invocations and decision accuracy (in parentheses) of different methods on MetaTool.}
\label{tab:num_tool_invocations}
\end{table}

Secondly, we evaluate the latency overhead introduced by \mname on \dname-RAG. This experiment adopts a standard RAG pipeline involving a retrieval tool, consisting of the following stages:
\begin{itemize}[leftmargin=*]
    \item \textbf{Pre-processing:} Queries are rewritten to be clearer and more specific for search. If the query is complex, it is decomposed into simpler, related sub-queries.
    \item \textbf{Search Engine:} Google Search is used to retrieve the top-5 most relevant web pages.
    \item \textbf{Post-processing:} The model integrates the original query with the retrieved content to generate an answer. When needed, relevant details are summarized or extracted to directly address the user's query.
\end{itemize}

\begin{table}[t]
\small
\centering
\begin{tabular}{lccc}
    \toprule
    \textbf{Method} &\textbf{Search Overhead (s)} &\textbf{w/ ctx} & \textbf{w/o ctx} \\
    \midrule
    Always Call          & 14.8 & 0.00 & 0.00 \\
    Naive                & 14.8 & 0.94 & 0.85 \\
    $P_{\text{Yes}}$     & 14.8 & 0.98 & 0.89 \\
    \mname               & 14.8 & 1.74 & 0.95 \\
    \bottomrule
\end{tabular}
\caption{Increased latency (in seconds) introduced by different methods when deciding whether to invoke a search on \dname-RAG. With context and without context are abbreviated as w/ cts and w/o ctx for brevity.}
\label{tab:increased_latency}
\end{table}

The average response time for invoking the retrieval tool (including pre-processing, search, and post-processing) is 14.8 seconds. While \mname introduces a slight latency overhead due to the computation of meta-cognition scores, this cost is minimal. As shown in Table~\ref{tab:increased_latency}, the additional latency incurred by \mname is negligible relative to the overall retrieval time, particularly for searches without context. For instance, \mname adds only 0.95 seconds (a 6.4\% increase over $P_{\text{Yes}}$), which is marginal compared to the 14.8-second baseline. More importantly, \mname avoids unnecessary tool invocations, particularly in multi-round retrieval settings, thereby improving overall response efficiency and mitigating the significant delays introduced by external tools.

\begin{table}[t]
\small
\renewcommand{\arraystretch}{1.3}
\resizebox{\linewidth}{!}{
    \centering
    \begin{tabular}{cl|cc|cc}
        \toprule
         \multicolumn{2}{c|}{ \multirow{3}{*}{\textbf{Method}}} & \multicolumn{2}{c|}{\textbf{Pre Fine-tuning}} & \multicolumn{2}{c}{\textbf{Post Fine-tuning}} \\
        \cmidrule(lr){3-4} \cmidrule(lr){5-6}
        & & w/ ctx & w/o ctx & w/ ctx & w/o ctx \\ 
        \midrule

        \parbox[t]{2.0mm}{\multirow{3}{*}{\rotatebox[origin=c]{90}{\textit{\textcolor{gray}{\textbf{LM3-8B}}}}} } 
        & Naive & 61.9 & 58.3 & 82.1 & 80.8 \\
        & $P_{\text{Yes}}$ & 63.5 & 62.7 & 81.7 & 80.8 \\
        & \mname & \textbf{65.0} & \textbf{74.0} & \textbf{84.3} & \textbf{82.3} \\
        \midrule
        \parbox[t]{2.0mm}{\multirow{3}{*}{\rotatebox[origin=c]{90}{\textit{\textcolor{gray}{\textbf{LM3-70B}}}}} }
        & Naive & 84.6 & 68.8 & 86.0 & 77.7 \\
        & $P_{\text{Yes}}$ & 84.8 & 73.7 & 86.2 & 77.1 \\
        & \mname & \textbf{85.4} & \textbf{79.6} & \textbf{87.3} & \textbf{81.2} \\
        \midrule
        \parbox[t]{2.0mm}{\multirow{3}{*}{\rotatebox[origin=c]{90}{\textit{\textcolor{gray}{\textbf{Mist-7B}}}}} }
        
        & Naive & 69.0 & 68.5 & 89.2 & 86.0 \\
        & $P_{\text{Yes}}$ & 71.2 & 73.1 & 89.2 & 85.0 \\
        & \mname & \textbf{75.4} & \textbf{74.7} & \textbf{90.2} & \textbf{86.5} \\
        \midrule
        \multicolumn{2}{c|}{\textbf{GPT-4-turbo}} &84.4 &61.3 & - & - \\
        \bottomrule
    \end{tabular}}
    \caption{Decision accuracy (\%) comparison between Naive, $P_{\text{Yes}}$ and \mname on the MetaTool benchmark. Note that we are unable to calculate $P_{\text{Yes}}$ or detect the internal states of proprietary LLMs such as GPT-4-turbo. As a result, GPT-4-turbo uses the Naive decision-making strategy.}
    \label{tab:restuls_on_MetaTool}
\end{table}

\begin{table*}[t]
    \centering
    \small
    \setlength{\tabcolsep}{3.8pt} 
    \renewcommand{\arraystretch}{1.45} 
    \begin{tabular}{cc|cc|cc|cc|cc|cc|cc}
        \toprule
        \multicolumn{2}{c|}{ \multirow{3}{*}{\textbf{Method}} }
        & \multicolumn{2}{c|}{\textbf{Task 1}} 
        & \multicolumn{2}{c|}{\textbf{Task 2}} 
        & \multicolumn{2}{c|}{\textbf{Task 3}} 
        & \multicolumn{2}{c|}{\textbf{Task 4}} 
        & \multicolumn{2}{c|}{\textbf{Task 5}} 
        & \multicolumn{2}{c}{\textbf{Task 6}} \\ 
        \cmidrule(lr){3-14} 
        & & w/ ctx & w/o ctx
        & w/ ctx& w/o ctx
        & w/ ctx& w/o ctx
        & w/ ctx& w/o ctx
        & w/ ctx& w/o ctx
        & w/ ctx& w/o ctx\\ 
        \midrule
        \parbox[t]{2.0mm}{\multirow{3}{*}{\rotatebox[origin=c]{90}{\textit{\textcolor{gray}{\textbf{LM3-8B}}}}} } 
        & Naive 
          & 70.0 & 65.0  
          & 62.3 & 80.3  
          & 54.3 & 78.7  
          & 66.0 & 50.0  
          & 70.5 & 54.0  
          & 60.5 & 53.5  
          \\
        & $P_{\text{Yes}}$   
          & 74.0 & 67.0  
          & 78.0 & 80.7  
          & 66.0 & \textbf{81.3}  
          & 66.0 & 62.0  
          & 72.0 & 71.5  
          & 62.0 & 64.5  
          \\
        & \textbf{MeCo}  
          & \textbf{79.0} & \textbf{72.0}  
          & \textbf{80.1} & \textbf{81.3}  
          & \textbf{73.3} & {79.5}  
          & \textbf{69.0} & \textbf{69.0}  
          & \textbf{74.0} & \textbf{78.5}  
          & \textbf{63.5} & \textbf{67.0}  
          \\
        \midrule
        \parbox[t]{2.0mm}{\multirow{3}{*}{\rotatebox[origin=c]{90}{\textit{\textcolor{gray}{\textbf{Mist-7B}}}}} }  
        & Naive 
          & 54.0 & 63.0  
          & 42.3 & 55.7  
          & 55.7 & 67.3  
          & 60.5 & 70.0  
          & 73.5 & 76.0  
          & 73.0 & 62.5  
          \\
        & $P_{\text{Yes}}$     
          & 54.0 & 63.0  
          & 45.0 & 60.0  
          & 56.7 & 70.7  
          & 66.5 & 71.0  
          & 73.0 & 76.0  
          & 73.5 & 63.0  
          \\
        & \textbf{MeCo}
          & \textbf{58.0} & \textbf{67.0}  
          & \textbf{66.7} & \textbf{66.0}  
          & \textbf{74.8} & \textbf{78.3}  
          & \textbf{69.0} & \textbf{78.5}  
          & \textbf{76.2} & \textbf{80.0}  
          & \textbf{74.0} & \textbf{65.5}  
          \\
        \bottomrule
    \end{tabular}
    \caption{Decision accuracy (\%) comparison between Naive, $P_{\text{Yes}}$, and \mname on \dname-Tool before fine-tuning. "With context" and "without context" are abbreviated as w/ ctx and w/o ctx, respectively.}
    \label{tab:meca_performance_pre}
\end{table*}

\begin{table*}[th]
    \centering
    \small
    \setlength{\tabcolsep}{3.8pt} 
    \renewcommand{\arraystretch}{1.45} 
    \begin{tabular}{cc|cc|cc|cc|cc|cc|cc}
        \toprule
        \multicolumn{2}{c|}{ \multirow{3}{*}{\textbf{Method}} }
        & \multicolumn{2}{c|}{\textbf{Task 1}} 
        & \multicolumn{2}{c|}{\textbf{Task 2}} 
        & \multicolumn{2}{c|}{\textbf{Task 3}} 
        & \multicolumn{2}{c|}{\textbf{Task 4}} 
        & \multicolumn{2}{c|}{\textbf{Task 5}} 
        & \multicolumn{2}{c}{\textbf{Task 6}} \\ 
        \cmidrule(lr){3-14} 
        & & w/ ctx & w/o ctx
        & w/ ctx& w/o ctx
        & w/ ctx& w/o ctx
        & w/ ctx& w/o ctx
        & w/ ctx& w/o ctx
        & w/ ctx& w/o ctx\\ 
        \midrule
        \parbox[t]{2.0mm}{\multirow{3}{*}{\rotatebox[origin=c]{90}{\textit{\textcolor{gray}{\textbf{LM3-8B-sft}}}}} } 
        & Naive & 69.0 & 80.0  
          & 53.3 & 61.0  
          & 59.0 & 68.7  
          & 74.0 & 77.0  
          & 71.0 & 78.5  
          & 78.5 & 83.0  
          \\
        & $P_{\text{Yes}}$     & \textbf{70.0} & 78.0  
          & 58.3 & 68.7  
          & 57.7 & 70.0  
          & 75.0 & 77.0  
          & \textbf{80.5} & \textbf{84.0}  
          & \textbf{81.5} & 82.0  
          \\
        & \textbf{MeCo}  & 69.0 & \textbf{80.0}  
          & \textbf{59.9} & \textbf{70.3}  
          & \textbf{60.0} & \textbf{73.4}  
          & \textbf{75.0} & \textbf{84.5}  
          & 79.5 & 82.0  
          & 80.0 & \textbf{86.5}  
          \\
        \midrule
        \parbox[t]{2.0mm}{\multirow{3}{*}{\rotatebox[origin=c]{90}{\textit{\textcolor{gray}{\textbf{Mist-7B-sft}}}}} }  
        & Naive & 68.0 & 64.0  
          & 52.3 & 53.0  
          & 58.3 & 73.7  
          & 92.5 & 77.5  
          & 87.5 & 82.0  
          & 85.0 & 70.5  
          \\
        & $P_{\text{Yes}}$     & 69.0 & 63.0  
          & 55.3 & 62.3  
          & 61.0 & 75.0  
          & 92.5 & 80.5  
          & 87.5 & \textbf{83.0}  
          & 86.5 & 78.0  
          \\
        & \textbf{MeCo}  & \textbf{71.0} & \textbf{66.0}  
          & \textbf{60.7} & \textbf{66.3}  
          & \textbf{65.7} & \textbf{82.0}  
          & \textbf{95.0} & \textbf{87.0}  
          & \textbf{88.0} & 82.0  
          & \textbf{88.0} & \textbf{80.5}  
          \\
        \bottomrule
    \end{tabular}
    \caption{Decision accuracy (\%) comparison between Naive, $P_{\text{Yes}}$ and \mname on \dname-Tool after fine-tuning on the crafted SFT data. "With context" and "without context" are abbreviated as w/ ctx and w/o ctx, respectively.}
    \label{tab:meca_performance_post}
\end{table*}

\subsection{\mname in Adaptive Tool Use}
In this experiment, we sampled a subset of queries from the MetaTool benchmark to construct validation data for determining the optimal thresholds for $P_{\text{Yes}}$ and \mname. These thresholds were then applied to the test queries in both MetaTool and \dname-Tool (Task 1 and 4). Given the fundamental differences between the queries in MetaTool and those in Task 2, 3, 5, and 6 of \dname-Tool, we randomly sampled 100 queries from each of these categories to serve as held-out test data. The thresholds for both $P_{\text{Yes}}$ and \mname were fit using the remaining data in each respective category. Comprehensive evaluation results are reported in Table \ref{tab:restuls_on_MetaTool}, Table \ref{tab:meca_performance_pre}, and Table \ref{tab:meca_performance_post}. We report the decision accuracy of GPT-4-turbo as a reference point to illustrate the difficulty level of the queries in the MetaTool benchmark. We highlight two key observations:

\textbf{1. Superiority of \mname:} Across both benchmarks, \mname significantly improves the model’s decision accuracy regarding tool use, outperforming the $P_{\text{Yes}}$ baseline by a notable margin. This highlights the effectiveness of the meta-cognition-based trigger mechanism. Crucially, \mname's advantages are consistent across different backbone models and evaluation settings—both with and without context, as well as before and after fine-tuning.

Importantly, the gains from \mname come at minimal cost: it is a fine-tuning-free, easily integrable module. Note that fine-tuning and \mname are two orthogonal approaches, and \mname can provide additional benefits to fine-tuned models. Notably, fine-tuned models often struggle to generalize to "out-of-distribution" scenarios. For instance, we observed a performance drop in the fine-tuned \texttt{LM3-8B} on Tasks 2 and 3 of \dname-Tool. In contrast, \mname's improvements are robust and consistently maintained across varied testing conditions.

The strong performance of \mname on \dname is particularly noteworthy, as \dname presents more complex and realistic queries and user–assistant interactions that better reflect real-world usage. This underscores \mname’s potential for practical deployment and its effectiveness in diverse and realistic settings.

\textbf{2. Transferability:} Results from Tasks 1 and 4 in Table~\ref{tab:meca_performance_pre} and Table~\ref{tab:meca_performance_post} show that both $P_{\text{Yes}}$ and \mname, when fitted on one benchmark, can generalize effectively to others. Despite the differences in tool sources and query styles between MetaTool and \dname benchmarks, both methods demonstrate promising cross-benchmark performance. We hypothesize that an LLM’s internal meta-cognition is primarily model-dependent, and once a decision strategy (\ie thresholds) is learned on one dataset, it can generalize to others. Although aligning the decision strategy with the target data is always preferable, \mname performs robustly even in direct transfer settings, highlighting its robustness and adaptability.

\begin{table}[ht]
\centering
\begin{tabular}{clc}
    \toprule
    {\textbf{Model}} & \textbf{Method} & \textbf{Accuracy (\%)} \\
    \midrule
    \multirow{3}{*}{\textbf{LM3-8B}} & Naive & 63.0 \\
    & $P_{\text{Yes}}$ & 75.0 \\
    & \mname & \textbf{76.0} \\
    \midrule
    \multirow{3}{*}{\textbf{Mist-7B}} & Naive & 84.0 \\
    & $P_{\text{Yes}}$ & 84.0 \\
    & \mname & \textbf{86.0} \\
    \midrule
    \textbf{GPT-4-Turbo} & Naive & 84.0 \\
    \bottomrule
\end{tabular}
\caption{Decision accuracy (\%) comparison between Naive, $P_{\text{Yes}}$ and \mname on \dname-RAG.}
\label{tab:rag_on_meca}
\end{table}

\subsection{\mname in Adaptive RAG}
We further evaluate the effectiveness of \mname in the adaptive RAG task, where LLMs must decide whether to retrieve external information to accurately answer a user query. In line with standard settings for adaptive RAG, no explicit reasons or examples are provided in the prompts—only the raw query is given. To contextualize the difficulty of the queries in \dname-RAG, we report the decision accuracy of GPT-4-turbo as a reference. As shown in Table~\ref{tab:rag_on_meca}, \mname consistently demonstrates robust performance across multiple models and outperforms the baseline methods, validating its utility as a general-purpose trigger mechanism beyond tool use.

\section{Conclusion}
We introduce the concept of adaptive tool use, which advances traditional tool-learning paradigms that often invoke external tools indiscriminately. We propose \mname, a lightweight and fine-tuning-free plug-in module that leverages meta-cognitive signals within LLMs to make more informed decisions about tool invocation. \mname uses a probing mechanism to assess internal representations, enabling more precise determinations of when external tools and retrieval are truly necessary. To support rigorous evaluation, we introduce \dname, a new benchmark that captures both tool-use and retrieval-awareness across realistic and diverse scenarios. Through extensive experiments on both the MetaTool and \dname benchmarks, we demonstrate that \mname significantly enhances decision accuracy regarding the necessity of external tools and retrieval, across multiple models and settings. Our findings suggest that incorporating meta-cognition into the LLM tool-use framework can yield more efficient and reliable decision-making in practical applications.

\section{Limitations}
Our current evaluation of the \dname benchmark primarily focuses on the decision-making component—whether tool use or retrieval is necessary—rather than the full end-to-end performance of LLMs in completing tasks using tools. This includes downstream tasks such as selecting the correct tool, filling in parameters appropriately, and generating accurate final responses. Evaluating these stages would require substantial human annotation effort and introduces complexity beyond the current scope. While \mname could be extended to assist in downstream tasks such as parameter filling, doing so would increase inference latency. Exploring how to effectively integrate \mname into the full tool-use pipeline—balancing accuracy, efficiency, and usability—remains an important avenue for future research.

\bibliography{custom}

\clearpage
\appendix
\input{appendix}

\end{document}

%% file: appendix.tex
\onecolumn
\section{\dname Statistics}\label{app:sec:meca}

\subsection{\dname Statistics}
Table~\ref{tab:MeCa} summarizes the statistics of \dname. In Tasks 1 and 4, positive queries require a specific external tool to address the user queries, while negative queries can be answered using the LLM's internal capabilities alone. In Tasks 2 and 5, a tool name and its description are provided alongside the user query; the LLM must determine whether this specific tool is necessary to resolve the query. Neutral queries in Tasks 2 and 3 indicate that external tools are needed, but the given tool is irrelevant to solving the query. In Tasks 3 and 6, a list of tools (ranging from two to five) is provided with the user query. For multi-turn queries, a dialogue between the user and the assistant is presented, and the assistant must decide whether to rely on external tools to respond to the final turn.

\begin{table}[ht]
\centering
\begin{tabular}{cll}
\toprule
\textbf{Task} & \textbf{Category} & \textbf{Count} \\
\midrule
\multirow{2}{*}{\dname-Tool-Task1} & Positive queries without tools & 500 \\
                       & Negative queries without tools & 500 \\
\midrule
\multirow{3}{*}{\dname-Tool-Task2} & Positive queries with relevant tools & 500 \\
                       & Negative queries with tools & 500 \\
                       & Neutral queries with irrelevant tools & 500 \\
\midrule
\multirow{3}{*}{\dname-Tool-Task3} & Positive queries with a tool list & 500 \\
                       & Negative queries with a tool list & 500 \\
                       & Neutral queries with a tool list & 500 \\
\midrule
\multirow{2}{*}{\dname-Tool-Task4} & Multi-turn Negative queries without tools & 500 \\
                       & Multi-turn Positive queries without tools & 500 \\
\midrule
\multirow{2}{*}{\dname-Tool-Task5} & Multi-turn Positive queries with relevant tools & 500 \\
                       & Multi-turn Negative queries with tools & 500 \\
\midrule
\multirow{2}{*}{\dname-Tool-Task6} & Multi-turn Positive queries with a tool list & 500 \\ 
                       & Multi-turn Negative queries with a tool list & 500 \\
\midrule
\multirow{2}{*}{\dname-RAG} & Positive RAG & 150 \\
                       & Negative RAG & 150 \\
\bottomrule
\end{tabular}\caption{Tool use categories and counts.}
\label{tab:MeCa}
\end{table}

We directly transfer the $l_{yes}$ and $l_{no}$ thresholds of \mname, fitted on the MetaTool dataset, to Tasks 1 and 4 in \dname-Tool, with results shown in Table~\ref{tab:meca_performance_pre} and Table~\ref{tab:meca_performance_post}. Since the remaining tasks in \dname-Tool (Tasks 2, 3, 5, and 6) differ significantly from MetaTool and involve more complex user queries, we randomly sample 100 queries from each of these tasks as hold-out test data. The remaining data is used to fit the thresholds for $P_{\text{yes}}$ and \mname.

\subsection{\dname Creation}
\paragraph{\dname-Tool} To construct the \dname-Tool dataset, we followed a meticulous and structured process to ensure the queries are aligned with current LLM capabilities:
\begin{enumerate}[leftmargin=*, itemsep=0em, topsep=-0.0em]
    \item \textbf{Collection of diverse scenarios:} We began by gathering a wide range of domains and conversational scenarios from various online corpora. This step ensured that the subsequent synthetic APIs and conversations were grounded in realistic and diverse contexts.
    \item \textbf{Synthetic API design:} Based on the collected scenarios, we synthetically created 500 distinct APIs, emulating patterns from real-world applications to span multiple domains.
    \item \textbf{Query generation:} For each query, APIs were randomly sampled from the synthetic API pool. User queries were then constructed to reflect one of the following types: (i) Queries that require invoking the provided APIs; (ii) Queries solvable without external tools, relying solely on the LLM’s internal knowledge; or (iii) Queries where the provided APIs do not include the necessary tools to directly answer the query.
    \item \textbf{Human verification:} All generated queries underwent a rigorous human review process to ensure validity, correctness, and proper categorization. This step guaranteed high-quality annotations and alignment with task definitions.
\end{enumerate}

\paragraph{\dname-RAG} The dataset was constructed as follows: we selected a subset of fact-based entries from the RepE dataset~\citep{zou2023representation}, which contains well-known facts such as "The Earth orbits the Sun." These facts were used as target responses, and the leading proprietary LLM (\ie GPT-4-turbo) was prompted to generate corresponding user queries. Since these queries involve widely known information already embedded in LLMs, they do not require retrieval and thus serve as negative RAG examples. For positive RAG examples, we scraped recent news articles from the past few months, ensuring that the content was unlikely to have been seen during LLM training. Using a similar prompting process, we generated user queries based on this up-to-date information. These queries require retrieval, as they concern facts that are either unknown to or not yet encoded in the model’s training data. The overall distribution of \dname is illustrated in Figure~\ref{fig:MeCa}.

\section{Related Work}
\paragraph{Tool Use in LLMs}
LLMs have progressed from understanding and generating human-like text to utilizing external tools based on natural language instructions. This evolution expands their application beyond basic conversational tasks to enable dynamic interactions across diverse functional domains, such as facility management and professional services~\citep{patil2023gorilla, liu2023bolaa,qin2023toolllm,chen2023agentverse,zhang2025correct}. For example, Toolformer~\citep{schick2024toolformer} enables LLMs to use external tools via simple APIs through a supervised fine-tuning (SFT) model. \cite{liuapigen} demonstrates strong executable functional API calls across different domains. ToolACE~\citep{liu2024toolace} trained on synthesized data, achieves state-of-the-art results on the Berkeley Function-Calling Leaderboard~\citep{berkeley-function-calling-leaderboard}, even with a relatively small model size of 8B parameters. Despite their growing popularity and capabilities, tool use in LLMs often depends on strategies like verbal feedback, which are hampered by the quality of the datasets used for fine-tuning. Several benchmarks/datasets have been developed to support tool use in a data-centric way, such as API-Bank~\citep{li2023comprehensive}, which provides a set of tool-use dialogues with various APIs to assess the LLM's tool use capabilities, Toolalpaca~\citep{tang2023toolalpaca} constructs a comprehensive tool-use corpus derived from collected real-world APIs, designed specifically to fine-tune LLMs for better tool utilization. ToolBench~\citep{qin2023toolllm} focuses on creating a synthetic instruction-tuning dataset for tool use. However, these methods rely solely on superficial textual information, without probing deeper into the LLM’s internal states to explain or justify when and why a tool should be called, resulting in an inability to accurately determine the optimal timing for tool invocation.

\paragraph{Adaptive RAG}
RAG has shown success in supporting AI systems that require up-to-date information or access domain-specific knowledge, particularly where the scope of queries is not seen in the training data of LLMs~\citep{lewis2020retrieval, ren2023investigating,vu2023freshllms,izacard2023atlas,dong2025benchmarking}. This paper is also consistent with the trend towards the adaptive RAG paradigm, which is designed to assess whether a query can be directly answered by the LLMs or requires external data retrieval~\citep{asai2023self, jiang2023active}. Specifically, a simple query within the LLM's knowledge should be directly answered by the LLMs themselves. On the other hand, for complex queries or questions about data they have not been trained on, RAG intervenes to prevent incorrect out-of-date answers or hallucination~\citep {JiSurvey}. This mechanism allows RAG to dynamically adjust operational strategies of retrieval-augmented LLMs by assessing the boundary of LLM's self-knowledge and the complexity of the query, thereby minimizing unnecessary computational overhead when the queries are answerable by LLMs themselves. Similar to the LLMs function-calling, the decision of retrieval timing typically hinges on three primary methods: (i) explicit verbal feedback from LLMs~\citep{ding2024retrieve}, (ii) enhancements through fine-tuning~\citep{asai2023self}, or (iii) probability-based metrics~\citep{kadavath2022language,jiang2023active}. Specifically, \citet{he2021efficient} proposed enhancing the retrieval time efficiency by computing the probability of the next token via interpolating an LLM with a distribution calculated from the $k$ nearest context-token pairs. \citet{drozdov2022you} further extends $k$NN-LM to the adaptive paradigm by assigning the interpolation coefficient according to the retrieval quality measured by semantic similarity. \citet{asai2023self} introduces Self-RAG to improve generation quality and factuality by enabling adaptive retrieval and self-reflection. In contrast, this paper conceptualizes RAG as an external tool and highlights the importance of understanding the internal states of an LLM when developing the retrieval policy. 

\paragraph{Explainability of LLMs}
However, there is a considerable discrepancy between the LLM's decision mechanisms (often based on verbalized responses) and their internal cognition~\citep{zou2023representation}. The internal workings of LLMs are usually unclear, and this lack of transparency poses unwanted risks in downstream decision-making. Therefore, understanding and interpreting LLMs is crucial for elucidating their behaviors and limitations. To address this challenge, various explanations that provide insights into the inner workings of LLMs have been proposed~\citep{zhao2024explainability}: 
(i) Probing-based explanations: Probing uses vector representations to measure embedded knowledge~\citep{peters2018dissecting, jawahar} or examines specific knowledge during the LLM's generation process~\citep{li2022probing}, (ii) Neuron-level explanation: neuron analysis identifies critical neurons that are essential for model's performance~\citep{antverg2021pitfalls,bills2023language}, (iii) representation engineering (RepE): RepE leverages techniques inspired by cognitive neuroscience to identify and enhance the transparency of LLMs by uncovering their internal cognitive states~\citep{zou2023representation}. In this paper, we aim to detect the internal cognition of LLMs, and intervene in LLM's decisions, \ie ensuring more precise decisions on tool use and retrieval timing.

\section{Extended Results}\label{app:sec:extend_results}
\subsection{Prompting Strategies}

To determine the best prompting strategy for tool use, we explore five prompting strategies with multiple base models. The results are summarized in Table \ref{tab:prompting_strategies}.
\begin{enumerate} 
\item Yes/No + Explanation: The model first answers with "Yes" or "No" and then provides a brief explanation for its decision. 
\item Yes/No: The model answers solely with "Yes" or "No," without providing any explanation. 
\item No/Yes + Explanation: The model first answers with "No" or "Yes" and then provides a brief explanation for its decision. 
\item No/Yes: The model answers solely with "No" or "Yes," without providing any explanation. 
\item CoT (Chain of Thought): The model is instructed to think step-by-step, reasoning why it does or does not need external tools to address the user query, and finally concludes its decision with "Yes" or "No." 
\end{enumerate}

Note that there are no results for the CoT prompting strategy for the \texttt{Mistral-7b-instruct-v0.3} model. Regardless of the prompts used, the model consistently responds with "Yes/No" at the beginning, followed by an explanation of its decision. This behavior effectively mirrors the Yes/No+Explanation prompting strategy. Based on Table \ref{tab:prompting_strategies}, we make the following observations and provide corresponding analysis:
\begin{enumerate} 
\item Yes/No + Explanation generally performs the best out of the five prompting strategies. This approach provides a clear decision followed by reasoning, enhancing the model's reliability and user trust. 
\item CoT is not performing as well as expected. Through close human examination, we found that CoT results in long, complex answers where the model might ultimately conclude with a decision that contrasts with its prior reasoning process. This phenomenon is referred to as reasoning inconsistency, a challenge also reported in the literature~\citep{wei2022chain,lyu2023faithful}. Specifically, LLMs sometimes generate the correct answer following an invalid reasoning path or produce a wrong answer after a correct reasoning process, leading to inconsistency between the derived answer and the reasoning process. In contrast, the "Yes/No-Explanation" prompting strategy does not suffer from this reasoning inconsistency in our experiments, thereby achieving better performance compared to CoT.
\item Yes/No prompting strategy works better than No/Yes prompting. We hypothesize that this phenomenon is due to the data format in the pre-training data. For example, there are likely many more Yes/No answers and reasoning processes in the training data compared to No/Yes answers, influencing the model's performance. 
\end{enumerate}

\begin{tcolorbox}[colback=blue!6!white, colframe=black, title=Chain of Thought Prompting.]
You are an intelligent agent, and you need to constantly be aware of your own limitations. I will provide you with a user's query, and you should assess, based on your own capabilities, whether you need to use external tools to better address the user's query. Typically, there are four reasons why you might need to use external tools:\\
\begin{itemize}
    \item A. Solving issues with real-time or external data, databases, or APIs
    \item B. Handling specialized inputs/outputs
    \item C. Enhancing domain tasks beyond LLM's capabilities
    \item D. User customization, personalization, and interaction
\end{itemize}
Please think step by step, and provide a brief explanation for your decision at first. At last, please conclude with "Yes" if you need to use external tools, or "No" if you do not need external tools. \\

\textcolor{red}{\{Few-shot Examples\}}\\
User query: \textcolor{red}{\{query\}}\\
Answer: 
\end{tcolorbox}

We adopt \texttt{Llama-3-8b-instruct} and \texttt{Mistral-7b-instruct-v0.3} as our backbone models because they exhibit strong performance in adaptive tool use. We exclude \texttt{Llama-2-7b-chat} due to its poor performance and lack of discernment regarding the necessity of external tools. Additionally, we exclude \texttt{Llama-3.1-8b-instruct} as its performance is almost identical to that of \texttt{Llama-3-8b-instruct}.
\begin{table}[ht]
\centering
\begin{tabular}{clcccc}
    \toprule
    Model &Prompting Strategies & Accuracy & Precision & Recall & F1 Score \\
    \midrule
    \multirow{5}{*}{Llama-2-7b-chat} 
    & Yes/No+Explanation & 0.51 & 0.51 & 1.0 & 0.67 \\
    & Yes/No & 0.51 & 0.5 & 1.0 & 0.67 \\
    & No/Yes+Explanation & \textbf{0.52} & 0.51 & 1.0 & 0.67 \\
    & No/Yes & 0.51 & 0.5 & 1.0 & 0.67 \\
    & CoT & 0.51 & 0.5 & 0.99 & 0.67 \\
    \midrule
    \multirow{4}{*}{Llama-3-8b-instruct} 
    & Yes/No+Explanation & \textbf{0.72} & 0.82 & 0.57 & 0.67 \\
    & Yes/No & 0.63 & 0.61 & 0.72 & 0.66 \\
    & No/Yes+Explanation & 0.52 & 0.51 & 0.99 & 0.67 \\
    & No/Yes & 0.5 & 0.5 & 1.0 & 0.67 \\
    & CoT & 0.62 & 0.59 & 0.84 & 0.69 \\
    \midrule
    \multirow{4}{*}{Llama-3.1-8b-instruct} 
    & Yes/No+Explanation & \textbf{0.71} & 0.66 & 0.87 & 0.75 \\
    & Yes/No & 0.64 & 0.59 & 0.95 & 0.73 \\
    & No/Yes+Explanation & 0.57 & 0.54 & 0.97 & 0.69 \\
    & No/Yes & 0.53 & 0.51 & 0.99 & 0.68 \\
    & CoT & 0.63 & 0.62 & 0.91 & 0.71 \\
    \midrule
    \multirow{4}{*}{Mistral-7b-instruct-v0.3} 
    & Yes/No+Explanation & \textbf{0.74} & 0.68 & 0.89 & 0.77 \\
    & Yes/No & 0.70 & 0.64 & 0.92 & 0.75 \\
    & No/Yes+Explanation & 0.70 & 0.64 & 0.88 & 0.74 \\
    & No/Yes & 0.71 & 0.57 & 0.82 & 0.74 \\
    \bottomrule
\end{tabular}
\caption{Performance comparison of different prompting strategies.}
\label{tab:prompting_strategies}
\end{table}

\subsection{P(Yes) Approach}\label{app:subsec:p_yes}
The $P_{\text{Yes}}$ baseline introduces a \textit{Yes-score}, as defined in Section \ref{sec:setup}. This score provides a nuanced measure of the model’s confidence, refining the binary approach taken by the Naive baseline. The \textit{Yes-score} spans from $0$ to $1$, where a score of $0$ signifies a definite "No" and a score of $1$ signifies a definite "Yes". Scores close to $0.5$ reflect lower certainty in the model's response, signifying ambiguity in decision-making. By adjusting the model's output in cases where the \textit{Yes-score} is near $0.5$ to always "Yes/No" answer, we aim to enhance the accuracy of both tool use and RAG timing. We employ Equation (\ref{eq:p_yes_threshold}) to determine the optimal threshold $l$ for the \textit{Yes-score} based on training data, which is then applied to the test data.
\begin{equation}
\text{Decision} = 
\begin{cases} 
\text{Yes} & \text{if } \textit{Yes-score} > l \\
\text{No} & \text{if } \textit{Yes-score} \leq l 
\end{cases}
\label{eq:p_yes_threshold}
\end{equation}

\subsection{Distribution of P(Yes) and Meta-Cognition Scores}
Before delving into the analysis, we provide some background on the concept of calibration in the context of Large Language Models (LLMs). Calibration refers to the alignment between a model's predicted probabilities and the actual likelihood of those predictions being correct. A well-calibrated model generates probability scores that accurately reflect the true probability of its predictions.

\begin{figure}[ht]
    \centering
    \subfigure[Llama-3-8b with context]{
        \includegraphics[width=0.45\textwidth]{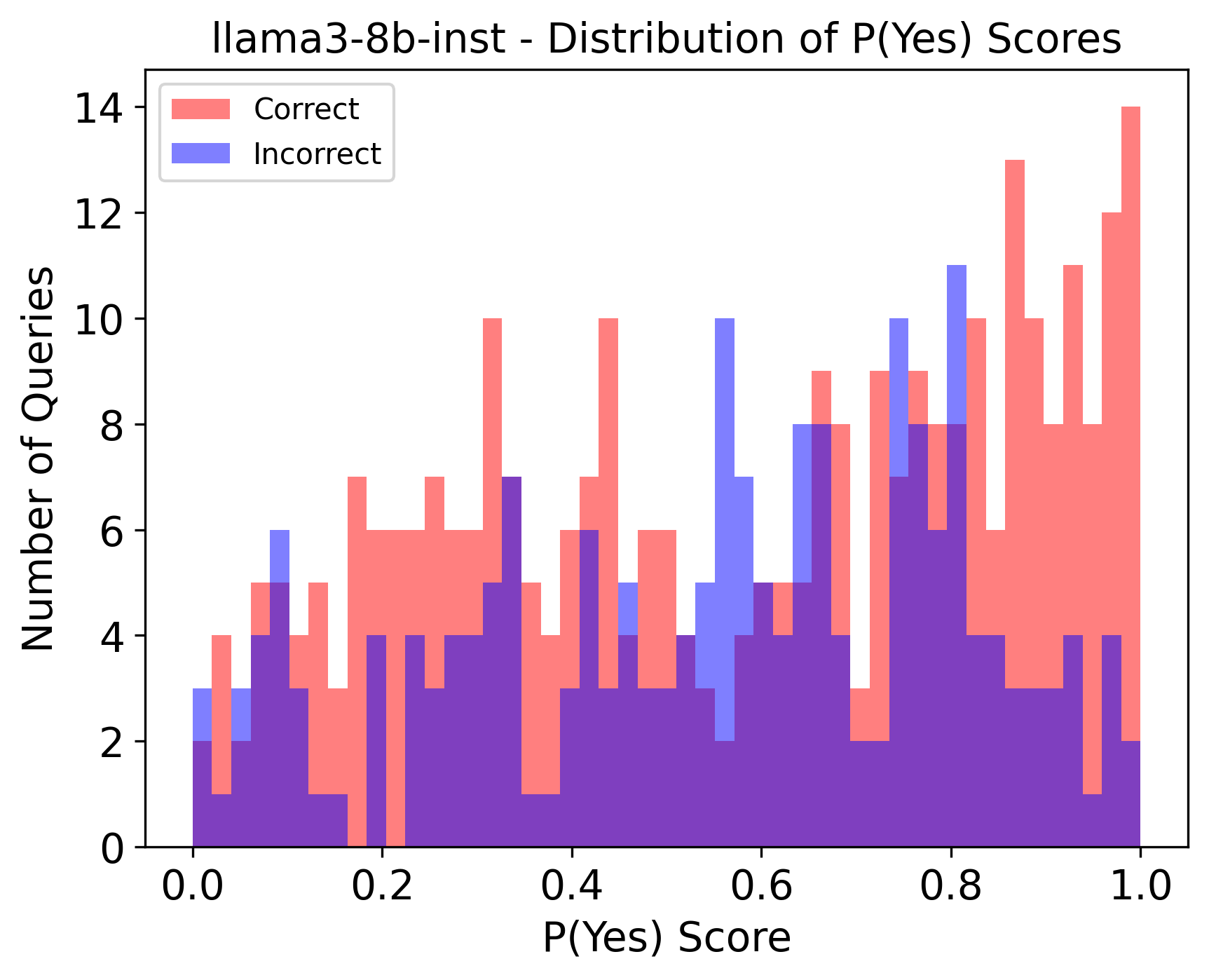}
    }
    \hfill
    \subfigure[Llama-3-8b without context]{
        \includegraphics[width=0.45\textwidth]{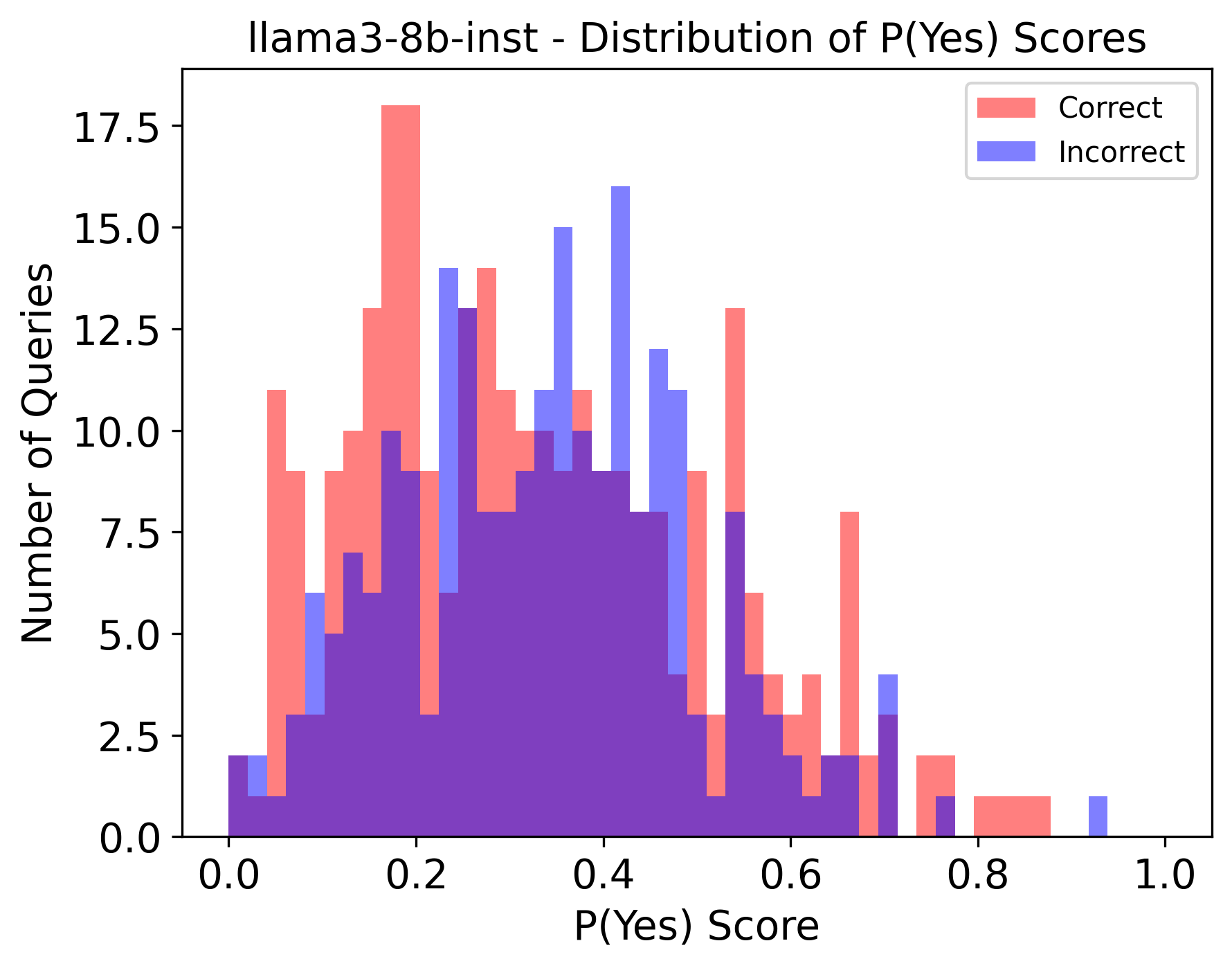}
    }
    \vskip\baselineskip
    \subfigure[Llama-3-8b-sft with context]{
        \includegraphics[width=0.45\textwidth]{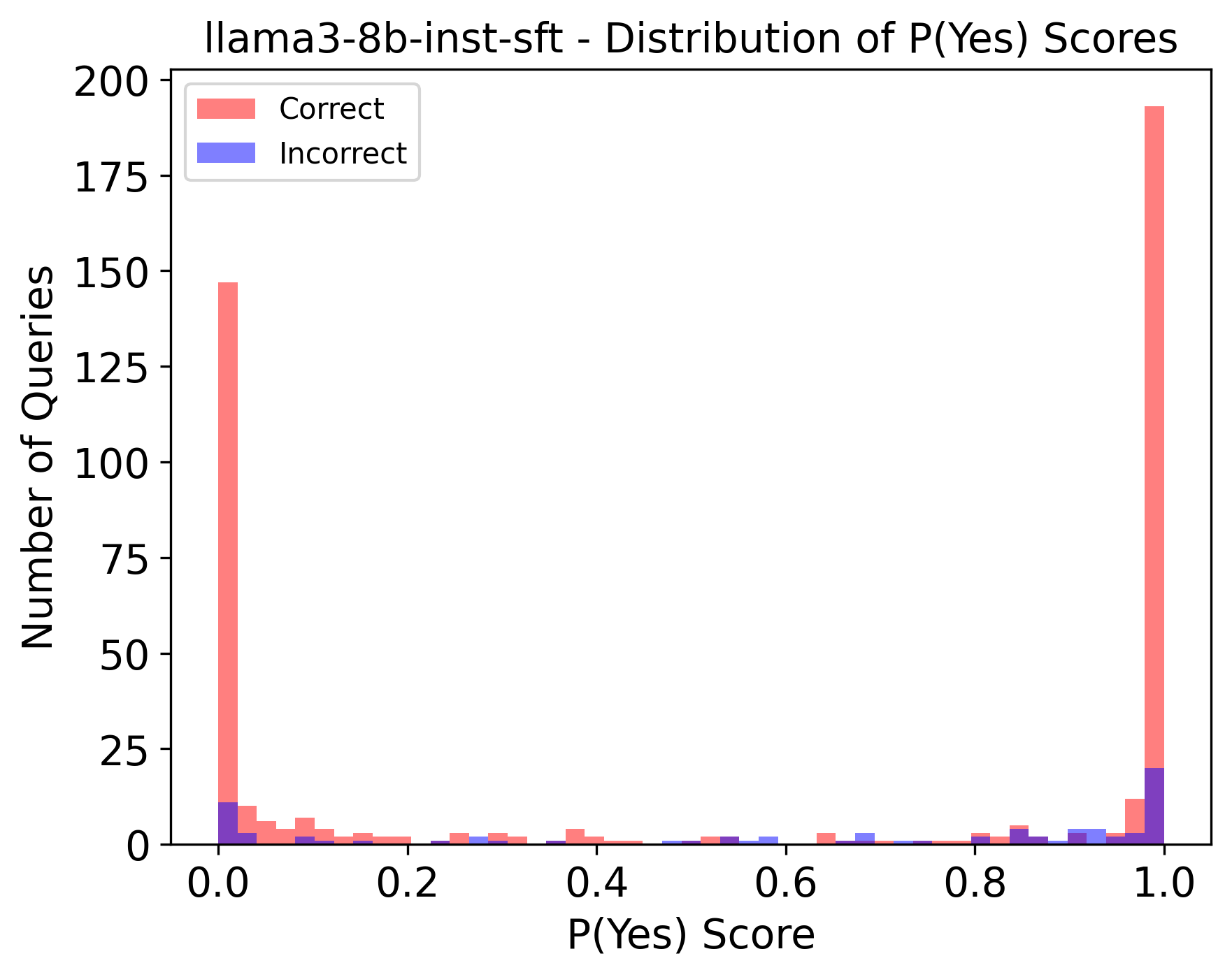}
    }
    \hfill
    \subfigure[Llama-3-8b-sft without context]{
        \includegraphics[width=0.45\textwidth]{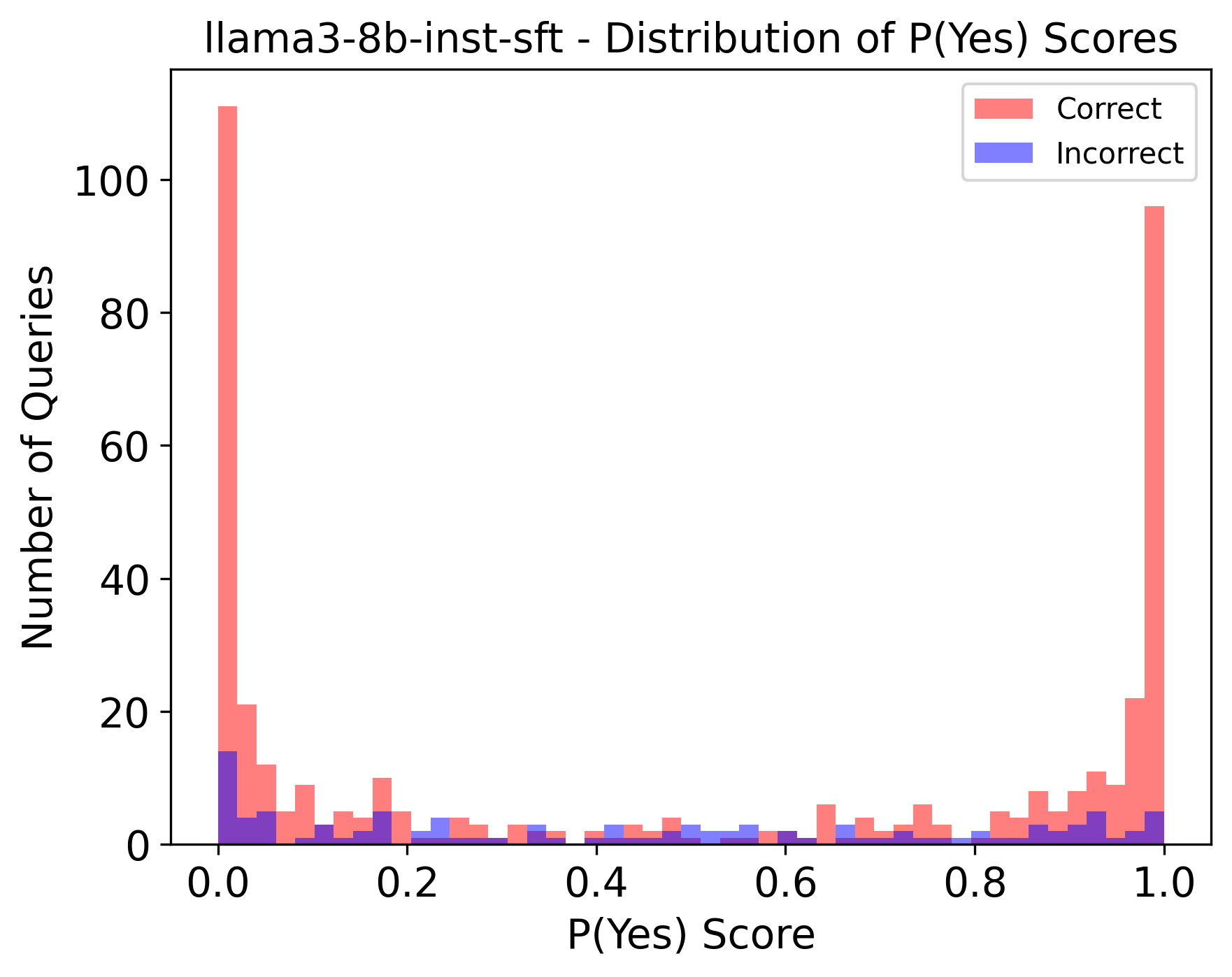}
    }
    \caption{Distribution of the $P_{\text{Yes}}$ scores of the correct Yes/No and incorrect Yes/No. Llama-3-8b is the model pre-fine-tuning and Llama-3-8b-sft is the model post-fine-tuning. Note that the scores are collected on the training data in the MetaTool benchmark.}
    \label{fig:p_yes_distribution}
\end{figure}

In Figure \ref{fig:p_yes_distribution}, we present the distribution of $P_{\text{Yes}}$ scores for both correct and incorrect Yes/No decisions. Our key observations are as follows:

\begin{enumerate} 
\item When the model is given detailed instructions and few-shot examples, it demonstrates poor calibration. As illustrated in Figure \ref{fig:p_yes_distribution}(a), the distributions of $P_{\text{Yes}}$ scores for correct and incorrect decisions do not show a clear distinction. 
\item Conversely, when the model lacks detailed context and must rely on its internal beliefs to make decisions, it exhibits improved calibration. In Figure \ref{fig:p_yes_distribution}(b), the peak of the distribution for correct scores clearly deviates from that of incorrect scores. 
\item After fine-tuning, the model displays significantly better calibration, as shown in Figures \ref{fig:p_yes_distribution}(c) and (d). Most correct decisions have $P_{\text{Yes}}$ scores of either 1 (indicating "Yes") or 0 (indicating "No"), while the $P_{\text{Yes}}$ scores for incorrect decisions vary between 0 and 1. 
\end{enumerate}

\subsection{Meta-Cognition Scores at Different Layers} 
\begin{figure}[ht]
  \centering
  \includegraphics[width=1.0\linewidth]{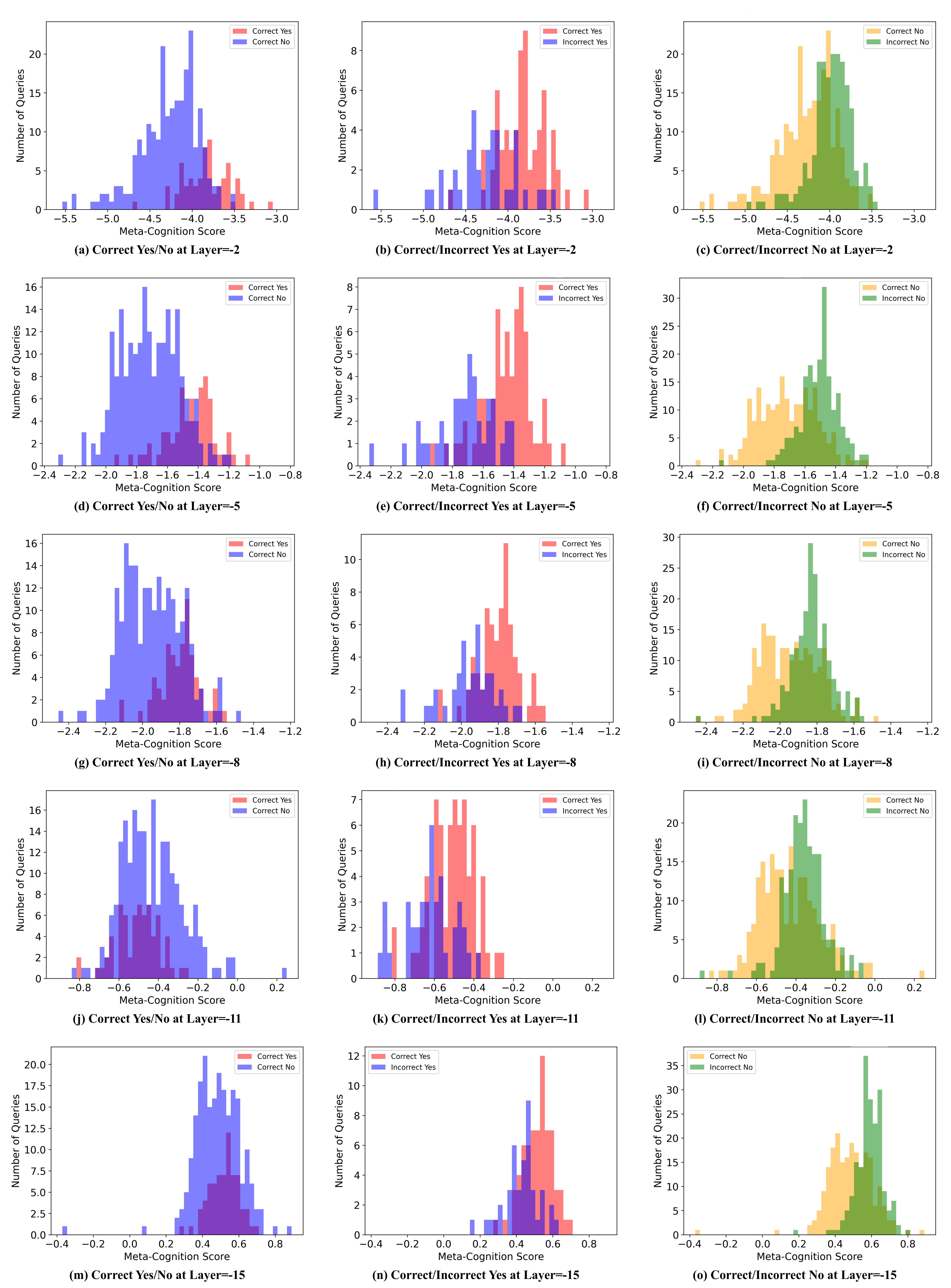}
  \caption{Distribution of meta-cognition scores for the first token at different layers. The results are collected using the Llama-3-8b model on the training data from the MetaTool benchmark.}
  \label{fig:metacog_different_layer}
\end{figure}

We examine the meta-cognition scores at various layers in the model and visualize the results in Figure \ref{fig:metacog_different_layer}. We focus on the meta-cognition scores at layers -2, -5, -8, -11, and -15 because these layers exhibit the highest classification accuracy, where layer -1 refers to the last layer before the output. Notably, the meta-cognition scores at different layers have distinct value ranges and slightly different distributions. Therefore, it is not reasonable to simply average the scores from different layers as the final score for a token, which has been a common approach in other research works based on RepE. In this study, we use the meta-cognition score from the second-to-last layer as the final score, as this layer demonstrates the highest classification accuracy and effectively differentiates between correct and incorrect responses.

\clearpage
\section{Probe Training}\label{app:sec:probe}
\subsection{Different Training Strategies}
Although it increases the length of the instructions and thus may degrade the signal we are detecting, we found that it is much better to provide the model with the query in the instruction than solely instruct the model to follow the ground truth explanations. Therefore, we include the queries in the contrastive instructions below. 
\begin{figure}[ht]
  \centering
  \includegraphics[width=1.0\linewidth]{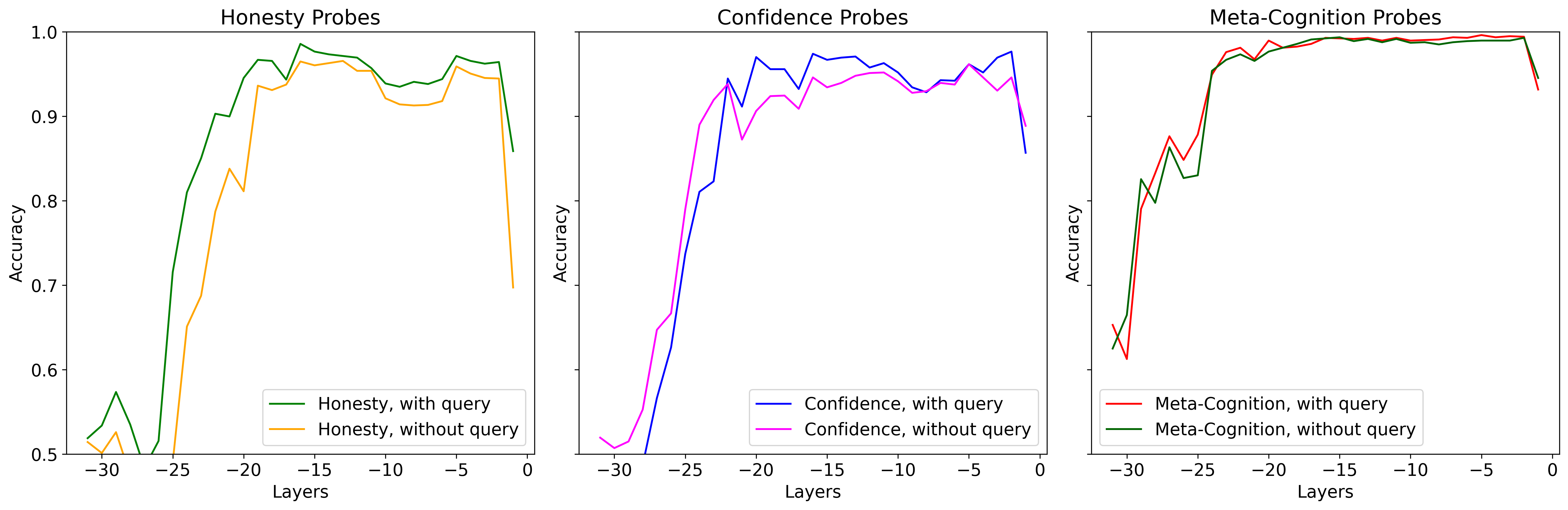}
  \caption{The classification accuracy of different probes trained with the query in the instruction and without the query in the instruction. Training data size is fixed as 2048 in this experiment.}
  \label{fig:probes_accuracy_dept_vs_indept}
\end{figure}

\subsection{Different Size of Training Data}
\begin{figure}[ht]
  \centering
  \includegraphics[width=1.0\linewidth]{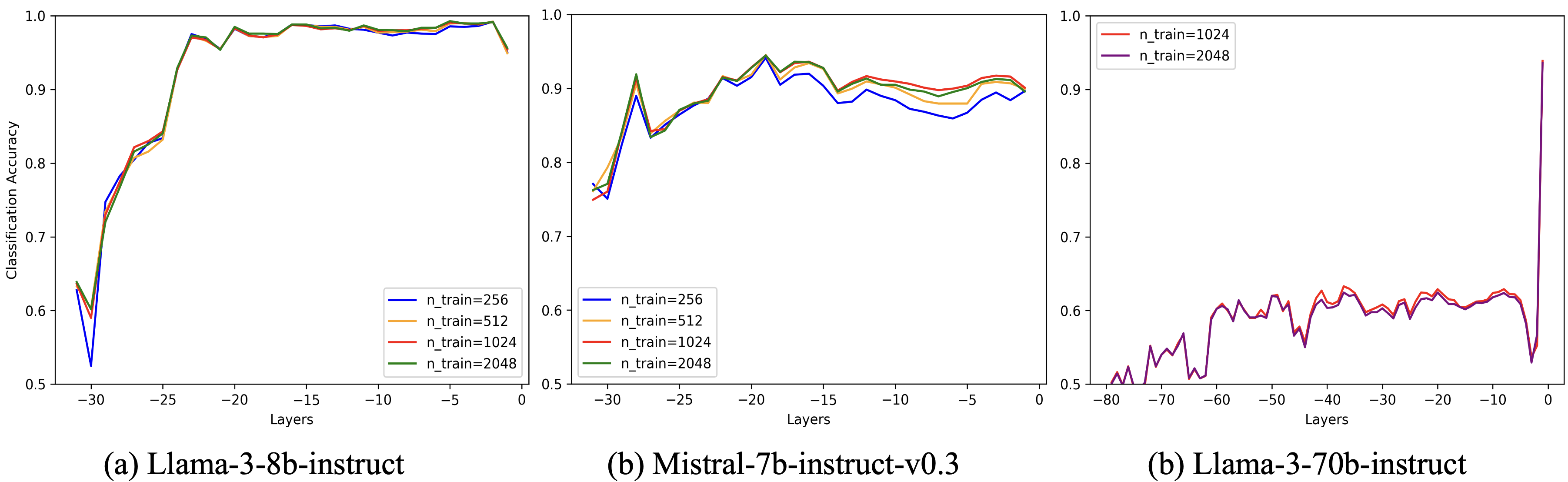}
  \caption{Training data size vs classification accuracy of meta-cognition probe in adaptive tool use.}
  \label{fig:probe_n_train}
\end{figure}

\begin{figure}[ht]
    \centering
    \subfigure[Llama-3-8b-instruct]{
        \includegraphics[width=0.48\textwidth]{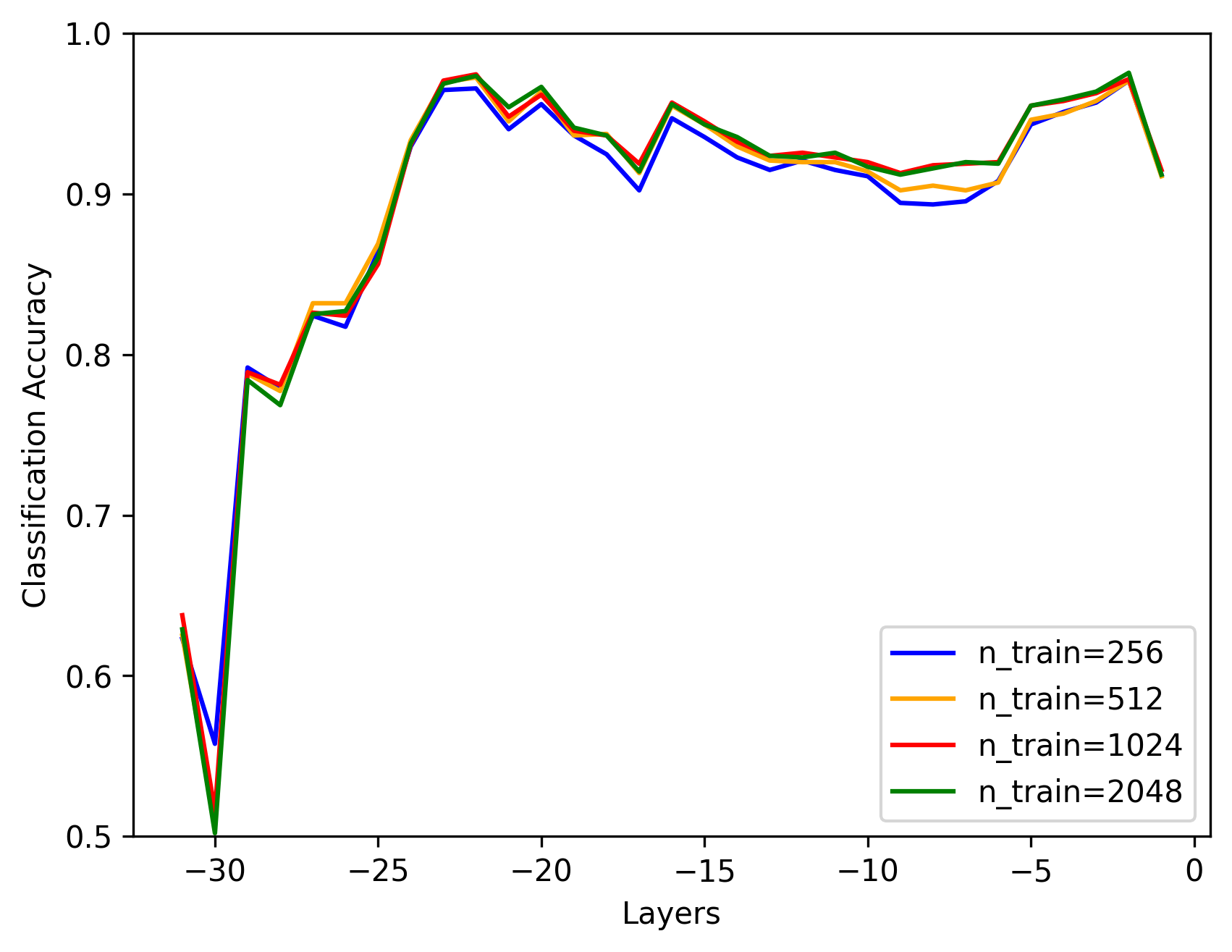}
    }
    \hfill
    \subfigure[Mistral-7b-instruct-v0.3]{
        \includegraphics[width=0.48\textwidth]{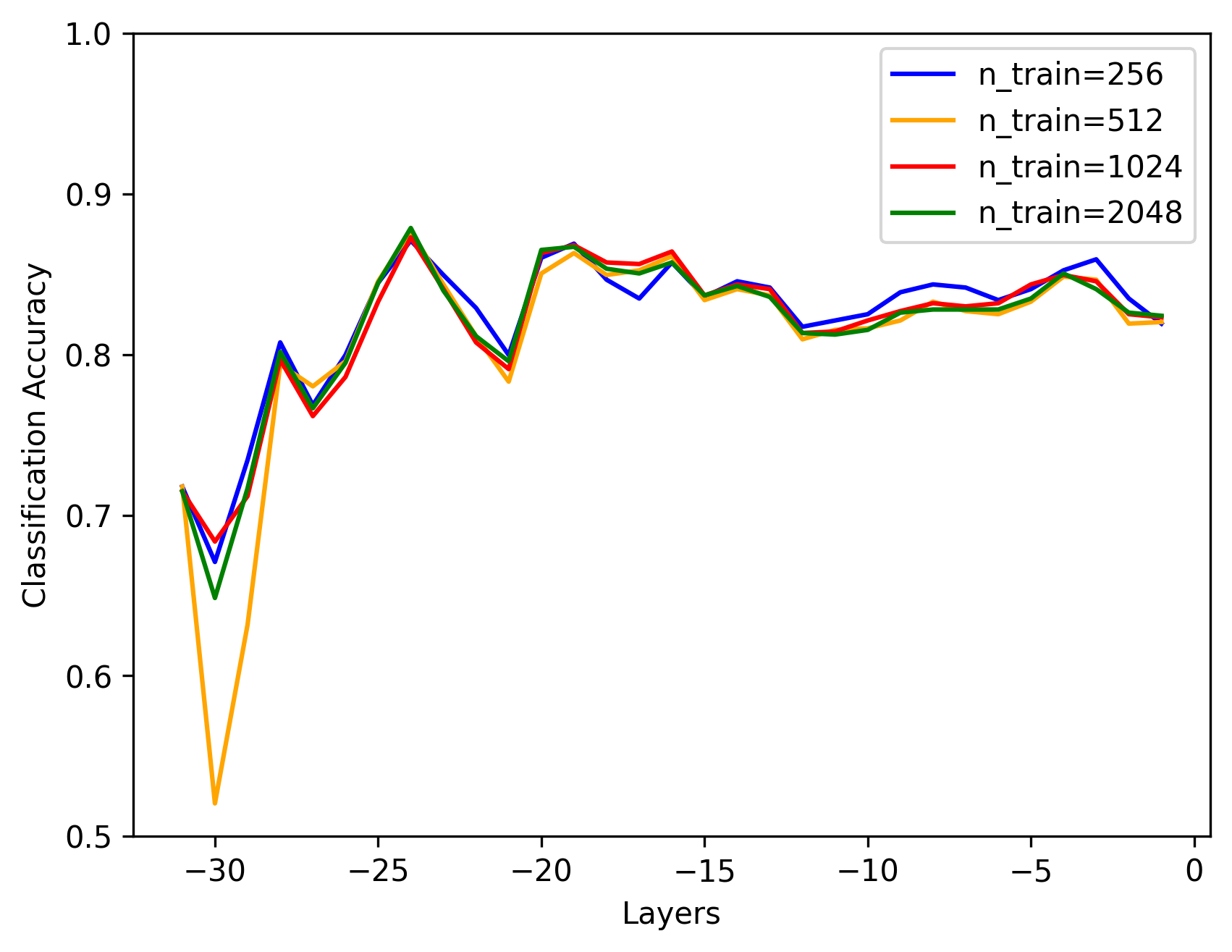}
    }
    \caption{Train data size vs classification accuracy of meta-cognition probe in adaptive RAG.}
    \label{fig:probe_n_train_rag}
\end{figure}

We further examine how the size of the training data affects the outcomes of the meta-cognition probe. Specifically, we analyze the performance of the trained probes with varying sizes of training data, as illustrated in Figure \ref{fig:probe_n_train} and Figure \ref{fig:probe_n_train_rag}. According to Equation (\ref{eq:RepE}), a sentence with 10 tokens can be used to create 10 training data pairs of experimental prompts and reference prompts. Typically, a brief explanation of why or why not to use external tools/RAG corresponds to around 30 to 50 tokens. Thus, a training data size of 256 requires fewer than 10 queries and their associated explanations.

Although different backbone models exhibit significantly varying classification accuracies—with Llama-3-8b achieving the highest and Llama-3-70b the lowest—we found that only a small amount of training data is sufficient to train a probe with near-optimal performance. We hypothesize that this is due to the linear nature of the PCA methods adopted in RepE.

\subsection{Contrastive Instructions for Various Probes}
We used the following instruction pair to collect contrastive data and train the honesty probe. Specifically, we instructed the model to be both honest and untruthful when explaining its reasoning for the necessity of tool use tes
\begin{tcolorbox}[colback=gray!10, colframe=black, title=Contrastive Instructions for training Honesty Probe]
\begin{tcolorbox}[colback=yellow!20, colframe=black, title=Honest Instruction]
Imagine you are an honest intelligent assistant explaining why you need or do not need to use an external tool to respond to the following user query.\\

User Query: \textcolor{red}{\{query\}}\\
Answer: 
\end{tcolorbox}

\begin{tcolorbox}[colback=green!20, colframe=black, title=Untruthful Instruction]
Imagine you are an untruthful intelligent assistant explaining why you need or do not need to use an external tool to respond to the following user query.\\

User Query: \textcolor{red}{\{query\}}\\
Answer: 
\end{tcolorbox}
\end{tcolorbox}

Similarly, we instructed the model to exhibit both confidence and unconfidence when we trained the confidence probe.
\begin{tcolorbox}[colback=gray!10, colframe=black, title=Contrastive Instructions for training Confidence Probe]
\begin{tcolorbox}[colback=yellow!20, colframe=black, title=Confident Instruction]
Imagine you are a confident intelligent assistant explaining why you need or do not need to use an external tool to respond to the following user query.\\

User Query: \textcolor{red}{\{query\}}\\
Answer: 
\end{tcolorbox}

\begin{tcolorbox}[colback=green!20, colframe=black, title=Unconfident Instruction]
Imagine you are an unconfident intelligent assistant explaining why you need or do not need to use an external tool to respond to the following user query.\\

User Query: \textcolor{red}{\{query\}}\\
Answer: 
\end{tcolorbox}
\end{tcolorbox}

For the meta-cognition probe, we instruct the model to exhibit strong meta-cognition by being constantly aware of its own limitations and capabilities and accurately assessing whether an external tool is necessary. Conversely, with weak meta-cognition, the model is often unaware of its own limitations and capabilities and struggles to assess the necessity of tool use.
\begin{tcolorbox}[colback=gray!10, colframe=black, title=Contrastive Instructions for training Meta-Cognition Probe]
\begin{tcolorbox}[colback=yellow!20, colframe=black, title=Strong Meta-Cognition Instruction in Adaptive Tool Use]
Imagine you are an intelligent assistant with strong meta-cognition, constantly aware of your own limitations and capabilities. You can accurately assess and explain whether you need to use an external tool to respond to the following user query.\\

User Query: \textcolor{red}{\{query\}}\\
Answer: 
\end{tcolorbox}

\begin{tcolorbox}[colback=green!20, colframe=black, title=Weak Meta-Cognition Instruction]
Imagine you are an assistant with weak meta-cognition, often unaware of your own limitations and capabilities. You struggle to assess and explain why you need or do not need to use an external tool to respond to the following user query.\\

User Query: \textcolor{red}{\{query\}}\\
Answer: 
\end{tcolorbox}
\end{tcolorbox}

The meta-cognition instruction for Adaptive RAG is similar to that in the adaptive tool use setting, with the only difference being that we replace the necessity of tool use with the necessity of RAG.
\begin{tcolorbox}[colback=gray!10, colframe=black, title=Contrastive Instructions for training Meta-Cognition Probe in Adaptive RAG]
\begin{tcolorbox}[colback=yellow!20, colframe=black, title=Strong Meta-Cognition Instruction]
Imagine you are an intelligent assistant with strong meta-cognition, constantly aware of your own limitations and capabilities. You can accurately assess and explain whether you need to perform Retrieval Augmented Generation (RAG) to respond to the following user query.\\

User Query: \textcolor{red}{\{query\}}\\
Answer: 
\end{tcolorbox}

\begin{tcolorbox}[colback=green!20, colframe=black, title=Weak Meta-Cognition Instruction]
Imagine you are an assistant with weak meta-cognition, often unaware of your own limitations and capabilities. You struggle to assess and explain why you need or do not need to perform Retrieval Augmented Generation (RAG) to respond to the following user query.\\

User Query: \textcolor{red}{\{query\}}\\
Answer: 
\end{tcolorbox}
\end{tcolorbox}

\clearpage
\section{Prompts}
\subsection{Prompts in Adaptive Tool Use}
We employ two types of prompts in our experiments: 1) prompts with context, which provide specific reasons for why LLMs may require external tools to complete user tasks. These prompts also include five randomly sampled examples to assist the model in making decisions; and 2) prompts without context, which are more concise and contain only the instruction and query. The exact prompts are provided below. Note that the example queries are randomly sampled in the MetaTool benchmark and we follow their setup and don't change the examples associated with queries. 

\begin{tcolorbox}[colback=blue!6!white, colframe=black, title=Prompt with context.]
You are an intelligent agent, and you need to constantly be aware of your own limitations. I will provide you with a user's query, and you should assess, based on your own capabilities, whether you need to use external tools to better address the user's query. Typically, there are four reasons why you might need to use external tools:
\begin{itemize}
    \item A. Solving issues with real-time or external data, databases, or APIs
    \item B. Handling specialized inputs/outputs
    \item C. Enhancing domain tasks beyond LLM's capabilities
    \item D. User customization, personalization, and interaction
\end{itemize}
If you think it's necessary to use external tools, please respond with "Yes"; otherwise, respond with "No". Additionally, you should provide a very brief explanation for your answer. Here are some examples:
\begin{itemize}
    \item Query: "Write an opinion piece about why diversity and inclusion is super important for the tech industry. The essay should be targeted at 'tech bros', and should avoid alienating them, but instead appeal to their logic; it should explain how diversity and inclusion of women, immigrants, etc. could benefit them specifically." Answer: No
    \item Query: "Are there any loopholes that hackers can exploit on my website?" Answer: Yes
    \item Query: "Plan a weekly lunch menu for a school. Write down a main dish, a carbohydrate side dish, a vegetable side dish, and a dessert for each day." Answer: No
    \item Query: "Can you break down the main points of this TED talk for me? Here's the YouTube link." Answer: Yes
    \item Query: "How's the weather in London right now?" Answer: No\\
\end{itemize}
User query: \textcolor{red}{\{query\}}\\
Answer: 
\end{tcolorbox}

\begin{tcolorbox}[colback=blue!6!white, colframe=black, title=Prompt without context.]
You are an intelligent agent, and you need to constantly be aware of your own limitations. I will provide you with a user's query, and you should assess, based on your own capabilities, whether you need to use external tools to better address the user's query. If you think it's necessary to use external tools, please respond with "Yes"; otherwise, respond with "No". Additionally, you should provide a very brief explanation for your answer.\\

User Query: \textcolor{red}{\{query\}}\\
Answer: 
\end{tcolorbox}

\subsection{Prompts in Adaptive RAG}
In adaptive RAG task, LLMs are typically not provided with any reasons or examples to help them make a decision. Following this setting, we conduct the experiments in adaptive RAG without providing context in the prompts as shown below.

\begin{tcolorbox}[colback=blue!6!white, colframe=black, title=Prompt without context.]
Imagine you are an intelligent assistant with strong meta-cognition, constantly aware of your own limitations and capabilities. Your task is to accurately assess and explain whether you need to perform Retrieval Augmented Generation (RAG) to respond to the following user query. If you determine that performing RAG is necessary, please respond with "Yes"; otherwise, respond with "No". Additionally, provide a very brief explanation for your decision.\\

User Query: \textcolor{red}{\{query\}}\\
Answer: 
\end{tcolorbox}

%% file: main.bbl
\begin{thebibliography}{48}
\providecommand{\natexlab}[1]{#1}

\bibitem[{Antverg and Belinkov(2021)}]{antverg2021pitfalls}
Omer Antverg and Yonatan Belinkov. 2021.
\newblock On the pitfalls of analyzing individual neurons in language models.
\newblock \emph{arXiv preprint arXiv:2110.07483}.

\bibitem[{Asai et~al.(2023)Asai, Wu, Wang, Sil, and Hajishirzi}]{asai2023self}
Akari Asai, Zeqiu Wu, Yizhong Wang, Avirup Sil, and Hannaneh Hajishirzi. 2023.
\newblock Self-rag: Learning to retrieve, generate, and critique through self-reflection.
\newblock \emph{arXiv preprint arXiv:2310.11511}.

\bibitem[{Bills et~al.(2023)Bills, Cammarata, Mossing, Tillman, Gao, Goh, Sutskever, Leike, Wu, and Saunders}]{bills2023language}
Steven Bills, Nick Cammarata, Dan Mossing, Henk Tillman, Leo Gao, Gabriel Goh, Ilya Sutskever, Jan Leike, Jeff Wu, and William Saunders. 2023.
\newblock Language models can explain neurons in language models.
\newblock \emph{URL https://openaipublic. blob. core. windows. net/neuron-explainer/paper/index. html.(Date accessed: 14.05. 2023)}, 2.

\bibitem[{Bricken et~al.(2023)Bricken, Templeton, Batson, Chen, Jermyn, Conerly, Turner, Anil, Denison, Askell, Lasenby, Wu, Kravec, Schiefer, Maxwell, Joseph, Hatfield-Dodds, Tamkin, Nguyen, McLean, Burke, Hume, Carter, Henighan, and Olah}]{bricken2023monosemanticity}
Trenton Bricken, Adly Templeton, Joshua Batson, Brian Chen, Adam Jermyn, Tom Conerly, Nick Turner, Cem Anil, Carson Denison, Amanda Askell, Robert Lasenby, Yifan Wu, Shauna Kravec, Nicholas Schiefer, Tim Maxwell, Nicholas Joseph, Zac Hatfield-Dodds, Alex Tamkin, Karina Nguyen, Brayden McLean, Josiah~E Burke, Tristan Hume, Shan Carter, Tom Henighan, and Christopher Olah. 2023.
\newblock \href {https://transformer-circuits.pub/2023/monosemantic-features/index.html} {Towards monosemanticity: Decomposing language models with dictionary learning}.
\newblock \emph{Transformer Circuits Thread}.

\bibitem[{Chen et~al.(2023)Chen, Su, Zuo, Yang, Yuan, Chan, Yu, Lu, Hung, Qian et~al.}]{chen2023agentverse}
Weize Chen, Yusheng Su, Jingwei Zuo, Cheng Yang, Chenfei Yuan, Chi-Min Chan, Heyang Yu, Yaxi Lu, Yi-Hsin Hung, Chen Qian, et~al. 2023.
\newblock Agentverse: Facilitating multi-agent collaboration and exploring emergent behaviors.
\newblock In \emph{The Twelfth International Conference on Learning Representations}.

\bibitem[{Ding et~al.(2024)Ding, Pang, Wei, Shen, and Cheng}]{ding2024retrieve}
Hanxing Ding, Liang Pang, Zihao Wei, Huawei Shen, and Xueqi Cheng. 2024.
\newblock Retrieve only when it needs: Adaptive retrieval augmentation for hallucination mitigation in large language models.
\newblock \emph{arXiv preprint arXiv:2402.10612}.

\bibitem[{Dong et~al.(2025)Dong, Chang, Huang, Wang, Tang, and Liu}]{dong2025benchmarking}
Kuicai Dong, Yujing Chang, Shijie Huang, Yasheng Wang, Ruiming Tang, and Yong Liu. 2025.
\newblock Benchmarking retrieval-augmented multimomal generation for document question answering.
\newblock \emph{arXiv preprint arXiv:2505.16470}.

\bibitem[{Drozdov et~al.(2022)Drozdov, Wang, Rahimi, Mccallum, Zamani, and Iyyer}]{drozdov2022you}
Andrew Drozdov, Shufan Wang, Razieh Rahimi, Andrew Mccallum, Hamed Zamani, and Mohit Iyyer. 2022.
\newblock You can’t pick your neighbors, or can you? when and how to rely on retrieval in the knn-lm.
\newblock In \emph{Findings of the Association for Computational Linguistics: EMNLP 2022}, pages 2997--3007.

\bibitem[{Gao et~al.(2023)Gao, Madaan, Zhou, Alon, Liu, Yang, Callan, and Neubig}]{gao2023pal}
Luyu Gao, Aman Madaan, Shuyan Zhou, Uri Alon, Pengfei Liu, Yiming Yang, Jamie Callan, and Graham Neubig. 2023.
\newblock Pal: Program-aided language models.
\newblock In \emph{International Conference on Machine Learning}, pages 10764--10799. PMLR.

\bibitem[{Hao et~al.(2024)Hao, Liu, Wang, and Hu}]{hao2024toolkengpt}
Shibo Hao, Tianyang Liu, Zhen Wang, and Zhiting Hu. 2024.
\newblock Toolkengpt: Augmenting frozen language models with massive tools via tool embeddings.
\newblock \emph{Advances in neural information processing systems}, 36.

\bibitem[{He et~al.(2021)He, Neubig, and Berg-Kirkpatrick}]{he2021efficient}
Junxian He, Graham Neubig, and Taylor Berg-Kirkpatrick. 2021.
\newblock Efficient nearest neighbor language models.
\newblock In \emph{Proceedings of the 2021 Conference on Empirical Methods in Natural Language Processing}, pages 5703--5714.

\bibitem[{He-Yueya et~al.(2023)He-Yueya, Poesia, Wang, and Goodman}]{he2023solving}
Joy He-Yueya, Gabriel Poesia, Rose~E Wang, and Noah~D Goodman. 2023.
\newblock Solving math word problems by combining language models with symbolic solvers.
\newblock \emph{arXiv preprint arXiv:2304.09102}.

\bibitem[{Huang et~al.(2023)Huang, Shi, Li, Fan, Wu, Zhang, Liu, Zhou, Wan, Gong, and Sun}]{huang2023MetaTool}
Yue Huang, Jiawen Shi, Yuan Li, Chenrui Fan, Siyuan Wu, Qihui Zhang, Yixin Liu, Pan Zhou, Yao Wan, Neil~Zhenqiang Gong, and Lichao Sun. 2023.
\newblock Metatool benchmark: Deciding whether to use tools and which to use.
\newblock \emph{arXiv preprint arXiv: 2310.03128}.

\bibitem[{Izacard et~al.(2023)Izacard, Lewis, Lomeli, Hosseini, Petroni, Schick, Dwivedi-Yu, Joulin, Riedel, and Grave}]{izacard2023atlas}
Gautier Izacard, Patrick Lewis, Maria Lomeli, Lucas Hosseini, Fabio Petroni, Timo Schick, Jane Dwivedi-Yu, Armand Joulin, Sebastian Riedel, and Edouard Grave. 2023.
\newblock Atlas: Few-shot learning with retrieval augmented language models.
\newblock \emph{Journal of Machine Learning Research}, 24(251):1--43.

\bibitem[{Jawahar et~al.(2019)Jawahar, Sagot, and Seddah}]{jawahar}
Ganesh Jawahar, Beno{\^i}t Sagot, and Djam{\'e} Seddah. 2019.
\newblock \href {https://inria.hal.science/hal-02131630} {{What does BERT learn about the structure of language?}}
\newblock In \emph{{ACL 2019 - 57th Annual Meeting of the Association for Computational Linguistics}}, Florence, Italy.

\bibitem[{Ji et~al.(2023)Ji, Lee, Frieske, Yu, Su, Xu, Ishii, Bang, Madotto, and Fung}]{JiSurvey}
Ziwei Ji, Nayeon Lee, Rita Frieske, Tiezheng Yu, Dan Su, Yan Xu, Etsuko Ishii, Ye~Jin Bang, Andrea Madotto, and Pascale Fung. 2023.
\newblock \href {https://doi.org/10.1145/3571730} {Survey of hallucination in natural language generation}.
\newblock \emph{ACM Comput. Surv.}, 55(12).

\bibitem[{Jiang et~al.(2023)Jiang, Xu, Gao, Sun, Liu, Dwivedi-Yu, Yang, Callan, and Neubig}]{jiang2023active}
Zhengbao Jiang, Frank~F Xu, Luyu Gao, Zhiqing Sun, Qian Liu, Jane Dwivedi-Yu, Yiming Yang, Jamie Callan, and Graham Neubig. 2023.
\newblock Active retrieval augmented generation.
\newblock \emph{arXiv preprint arXiv:2305.06983}.

\bibitem[{Kadavath et~al.(2022)Kadavath, Conerly, Askell, Henighan, Drain, Perez, Schiefer, Hatfield-Dodds, DasSarma, Tran-Johnson et~al.}]{kadavath2022language}
Saurav Kadavath, Tom Conerly, Amanda Askell, Tom Henighan, Dawn Drain, Ethan Perez, Nicholas Schiefer, Zac Hatfield-Dodds, Nova DasSarma, Eli Tran-Johnson, et~al. 2022.
\newblock Language models (mostly) know what they know.
\newblock \emph{arXiv preprint arXiv:2207.05221}.

\bibitem[{Komeili(2021)}]{komeili2021internet}
M~Komeili. 2021.
\newblock Internet-augmented dialogue generation.
\newblock \emph{arXiv preprint arXiv:2107.07566}.

\bibitem[{Levinstein and Herrmann(2024)}]{Levinstein_2024}
Benjamin~A. Levinstein and Daniel~A. Herrmann. 2024.
\newblock \href {https://doi.org/10.1007/s11098-023-02094-3} {Still no lie detector for language models: probing empirical and conceptual roadblocks}.
\newblock \emph{Philosophical Studies}.

\bibitem[{Lewis et~al.(2020)Lewis, Perez, Piktus, Petroni, Karpukhin, Goyal, K{\"u}ttler, Lewis, Yih, Rockt{\"a}schel et~al.}]{lewis2020retrieval}
Patrick Lewis, Ethan Perez, Aleksandra Piktus, Fabio Petroni, Vladimir Karpukhin, Naman Goyal, Heinrich K{\"u}ttler, Mike Lewis, Wen-tau Yih, Tim Rockt{\"a}schel, et~al. 2020.
\newblock Retrieval-augmented generation for knowledge-intensive nlp tasks.
\newblock \emph{Advances in Neural Information Processing Systems}, 33:9459--9474.

\bibitem[{Li et~al.(2022)Li, Cotterell, and Sachan}]{li2022probing}
Jiaoda Li, Ryan Cotterell, and Mrinmaya Sachan. 2022.
\newblock Probing via prompting.
\newblock \emph{arXiv preprint arXiv:2207.01736}.

\bibitem[{Li et~al.(2023)Li, Zhao, Yu, Song, Li, Yu, Li, Huang, and API-bank}]{li2023comprehensive}
Minghao Li, Yingxiu Zhao, Bowen Yu, Feifan Song, Hangyu Li, Haiyang Yu, Zhoujun Li, Fei Huang, and Yongbin~Li API-bank. 2023.
\newblock A comprehensive benchmark for tool-augmented llms.
\newblock In \emph{Proceedings of the 2023 Conference on Empirical Methods in Natural Language Processing}, pages 3102--3116.

\bibitem[{Liu et~al.(2024{\natexlab{a}})Liu, Zhang, Guo, Dong, Li, Lee, Zhang, and Liu}]{liu2024ctrla}
Huanshuo Liu, Hao Zhang, Zhijiang Guo, Kuicai Dong, Xiangyang Li, Yi~Quan Lee, Cong Zhang, and Yong Liu. 2024{\natexlab{a}}.
\newblock Ctrla: Adaptive retrieval-augmented generation via probe-guided control.
\newblock \emph{arXiv preprint arXiv:2405.18727}.

\bibitem[{Liu et~al.(2024{\natexlab{b}})Liu, Huang, Zeng, Hao, Yu, Li, Wang, Gan, Liu, Yu et~al.}]{liu2024toolace}
Weiwen Liu, Xu~Huang, Xingshan Zeng, Xinlong Hao, Shuai Yu, Dexun Li, Shuai Wang, Weinan Gan, Zhengying Liu, Yuanqing Yu, et~al. 2024{\natexlab{b}}.
\newblock Toolace: Winning the points of llm function calling.
\newblock \emph{arXiv preprint arXiv:2409.00920}.

\bibitem[{Liu et~al.(2023{\natexlab{a}})Liu, Wang, Wu, Li, Lv, Ling, Zhu, Zhang, Zheng, and Huang}]{liu2023aligning}
Wenhao Liu, Xiaohua Wang, Muling Wu, Tianlong Li, Changze Lv, Zixuan Ling, Jianhao Zhu, Cenyuan Zhang, Xiaoqing Zheng, and Xuanjing Huang. 2023{\natexlab{a}}.
\newblock Aligning large language models with human preferences through representation engineering.
\newblock \emph{arXiv preprint arXiv:2312.15997}.

\bibitem[{Liu et~al.(2023{\natexlab{b}})Liu, Yao, Zhang, Xue, Heinecke, Murthy, Feng, Chen, Niebles, Arpit et~al.}]{liu2023bolaa}
Zhiwei Liu, Weiran Yao, Jianguo Zhang, Le~Xue, Shelby Heinecke, Rithesh Murthy, Yihao Feng, Zeyuan Chen, Juan~Carlos Niebles, Devansh Arpit, et~al. 2023{\natexlab{b}}.
\newblock Bolaa: Benchmarking and orchestrating llm-augmented autonomous agents.
\newblock \emph{arXiv preprint arXiv:2308.05960}.

\bibitem[{Liu et~al.(2024{\natexlab{c}})Liu, Hoang, Zhang, Zhu, Lan, Kokane, Tan, Yao, Liu, Feng et~al.}]{liuapigen}
Zuxin Liu, Thai~Quoc Hoang, Jianguo Zhang, Ming Zhu, Tian Lan, Shirley Kokane, Juntao Tan, Weiran Yao, Zhiwei Liu, Yihao Feng, et~al. 2024{\natexlab{c}}.
\newblock Apigen: Automated pipeline for generating verifiable and diverse function-calling datasets.
\newblock In \emph{The Thirty-eight Conference on Neural Information Processing Systems Datasets and Benchmarks Track}.

\bibitem[{Lu et~al.(2024)Lu, Peng, Cheng, Galley, Chang, Wu, Zhu, and Gao}]{lu2024chameleon}
Pan Lu, Baolin Peng, Hao Cheng, Michel Galley, Kai-Wei Chang, Ying~Nian Wu, Song-Chun Zhu, and Jianfeng Gao. 2024.
\newblock Chameleon: Plug-and-play compositional reasoning with large language models.
\newblock \emph{Advances in Neural Information Processing Systems}, 36.

\bibitem[{Lyu et~al.(2023)Lyu, Havaldar, Stein, Zhang, Rao, Wong, Apidianaki, and Callison-Burch}]{lyu2023faithful}
Qing Lyu, Shreya Havaldar, Adam Stein, Li~Zhang, Delip Rao, Eric Wong, Marianna Apidianaki, and Chris Callison-Burch. 2023.
\newblock Faithful chain-of-thought reasoning.
\newblock In \emph{Proceedings of the 13th International Joint Conference on Natural Language Processing and the 3rd Conference of the Asia-Pacific Chapter of the Association for Computational Linguistics (Volume 1: Long Papers)}, pages 305--329.

\bibitem[{Ma{\'c}kiewicz and Ratajczak(1993)}]{mackiewicz1993principal}
Andrzej Ma{\'c}kiewicz and Waldemar Ratajczak. 1993.
\newblock Principal components analysis (pca).
\newblock \emph{Computers \& Geosciences}, 19(3):303--342.

\bibitem[{Patil et~al.(2023)Patil, Zhang, Wang, and Gonzalez}]{patil2023gorilla}
Shishir~G Patil, Tianjun Zhang, Xin Wang, and Joseph~E Gonzalez. 2023.
\newblock Gorilla: Large language model connected with massive apis.
\newblock \emph{arXiv preprint arXiv:2305.15334}.

\bibitem[{Peters et~al.(2018)Peters, Neumann, Zettlemoyer, and Yih}]{peters2018dissecting}
Matthew~E Peters, Mark Neumann, Luke Zettlemoyer, and Wen-tau Yih. 2018.
\newblock Dissecting contextual word embeddings: Architecture and representation.
\newblock \emph{arXiv preprint arXiv:1808.08949}.

\bibitem[{Qin et~al.(2023)Qin, Liang, Ye, Zhu, Yan, Lu, Lin, Cong, Tang, Qian et~al.}]{qin2023toolllm}
Yujia Qin, Shihao Liang, Yining Ye, Kunlun Zhu, Lan Yan, Yaxi Lu, Yankai Lin, Xin Cong, Xiangru Tang, Bill Qian, et~al. 2023.
\newblock Toolllm: Facilitating large language models to master 16000+ real-world apis.
\newblock \emph{arXiv preprint arXiv:2307.16789}.

\bibitem[{Qu et~al.(2024)Qu, Dai, Wei, Cai, Wang, Yin, Xu, and Wen}]{qu2024tool}
Changle Qu, Sunhao Dai, Xiaochi Wei, Hengyi Cai, Shuaiqiang Wang, Dawei Yin, Jun Xu, and Ji-Rong Wen. 2024.
\newblock Tool learning with large language models: A survey.
\newblock \emph{arXiv preprint arXiv:2405.17935}.

\bibitem[{Ren et~al.(2023)Ren, Wang, Qu, Zhao, Liu, Tian, Wu, Wen, and Wang}]{ren2023investigating}
Ruiyang Ren, Yuhao Wang, Yingqi Qu, Wayne~Xin Zhao, Jing Liu, Hao Tian, Hua Wu, Ji-Rong Wen, and Haifeng Wang. 2023.
\newblock Investigating the factual knowledge boundary of large language models with retrieval augmentation.
\newblock \emph{arXiv preprint arXiv:2307.11019}.

\bibitem[{Schick et~al.(2024)Schick, Dwivedi-Yu, Dess{\`\i}, Raileanu, Lomeli, Hambro, Zettlemoyer, Cancedda, and Scialom}]{schick2024toolformer}
Timo Schick, Jane Dwivedi-Yu, Roberto Dess{\`\i}, Roberta Raileanu, Maria Lomeli, Eric Hambro, Luke Zettlemoyer, Nicola Cancedda, and Thomas Scialom. 2024.
\newblock Toolformer: Language models can teach themselves to use tools.
\newblock \emph{Advances in Neural Information Processing Systems}, 36.

\bibitem[{Shen et~al.(2024)Shen, Song, Tan, Li, Lu, and Zhuang}]{shen2024hugginggpt}
Yongliang Shen, Kaitao Song, Xu~Tan, Dongsheng Li, Weiming Lu, and Yueting Zhuang. 2024.
\newblock Hugginggpt: Solving ai tasks with chatgpt and its friends in hugging face.
\newblock \emph{Advances in Neural Information Processing Systems}, 36.

\bibitem[{Tang et~al.(2023)Tang, Deng, Lin, Han, Liang, Cao, and Sun}]{tang2023toolalpaca}
Qiaoyu Tang, Ziliang Deng, Hongyu Lin, Xianpei Han, Qiao Liang, Boxi Cao, and Le~Sun. 2023.
\newblock Toolalpaca: Generalized tool learning for language models with 3000 simulated cases.
\newblock \emph{arXiv preprint arXiv:2306.05301}.

\bibitem[{Vu et~al.(2023)Vu, Iyyer, Wang, Constant, Wei, Wei, Tar, Sung, Zhou, Le et~al.}]{vu2023freshllms}
Tu~Vu, Mohit Iyyer, Xuezhi Wang, Noah Constant, Jerry Wei, Jason Wei, Chris Tar, Yun-Hsuan Sung, Denny Zhou, Quoc Le, et~al. 2023.
\newblock Freshllms: Refreshing large language models with search engine augmentation.
\newblock \emph{arXiv preprint arXiv:2310.03214}.

\bibitem[{Wang et~al.(2024)Wang, Ma, Feng, Zhang, Yang, Zhang, Chen, Tang, Chen, Lin et~al.}]{wang2024survey}
Lei Wang, Chen Ma, Xueyang Feng, Zeyu Zhang, Hao Yang, Jingsen Zhang, Zhiyuan Chen, Jiakai Tang, Xu~Chen, Yankai Lin, et~al. 2024.
\newblock A survey on large language model based autonomous agents.
\newblock \emph{Frontiers of Computer Science}, 18(6):186345.

\bibitem[{Wei et~al.(2022)Wei, Wang, Schuurmans, Bosma, Xia, Chi, Le, Zhou et~al.}]{wei2022chain}
Jason Wei, Xuezhi Wang, Dale Schuurmans, Maarten Bosma, Fei Xia, Ed~Chi, Quoc~V Le, Denny Zhou, et~al. 2022.
\newblock Chain-of-thought prompting elicits reasoning in large language models.
\newblock \emph{Advances in neural information processing systems}, 35:24824--24837.

\bibitem[{Wu et~al.(2024)Wu, Ahmad, Zhang, Ramanathan, and Ma}]{wurepoformer}
Di~Wu, Wasi~Uddin Ahmad, Dejiao Zhang, Murali~Krishna Ramanathan, and Xiaofei Ma. 2024.
\newblock Repoformer: Selective retrieval for repository-level code completion.
\newblock In \emph{Forty-first International Conference on Machine Learning}.

\bibitem[{Yan et~al.(2024)Yan, Mao, Ji, Zhang, Patil, Stoica, and Gonzalez}]{berkeley-function-calling-leaderboard}
Fanjia Yan, Huanzhi Mao, Charlie Cheng-Jie Ji, Tianjun Zhang, Shishir~G. Patil, Ion Stoica, and Joseph~E. Gonzalez. 2024.
\newblock Berkeley function calling leaderboard.
\newblock \url{https://gorilla.cs.berkeley.edu/blogs/8_berkeley_function_calling_leaderboard.html}.

\bibitem[{Yang et~al.(2023)Yang, Nachum, Du, Wei, Abbeel, and Schuurmans}]{yang2023foundation}
Sherry Yang, Ofir Nachum, Yilun Du, Jason Wei, Pieter Abbeel, and Dale Schuurmans. 2023.
\newblock Foundation models for decision making: Problems, methods, and opportunities.
\newblock \emph{arXiv preprint arXiv:2303.04129}.

\bibitem[{Zhang and Lee(2025)}]{zhang2025correct}
Delvin~Ce Zhang and Dongwon Lee. 2025.
\newblock Correct: Context-and reference-augmented reasoning and prompting for fact-checking.
\newblock \emph{arXiv preprint arXiv:2502.09635}.

\bibitem[{Zhao et~al.(2024)Zhao, Chen, Yang, Liu, Deng, Cai, Wang, Yin, and Du}]{zhao2024explainability}
Haiyan Zhao, Hanjie Chen, Fan Yang, Ninghao Liu, Huiqi Deng, Hengyi Cai, Shuaiqiang Wang, Dawei Yin, and Mengnan Du. 2024.
\newblock Explainability for large language models: A survey.
\newblock \emph{ACM Transactions on Intelligent Systems and Technology}, 15(2):1--38.

\bibitem[{Zou et~al.(2023)Zou, Phan, Chen, Campbell, Guo, Ren, Pan, Yin, Mazeika, Dombrowski et~al.}]{zou2023representation}
Andy Zou, Long Phan, Sarah Chen, James Campbell, Phillip Guo, Richard Ren, Alexander Pan, Xuwang Yin, Mantas Mazeika, Ann-Kathrin Dombrowski, et~al. 2023.
\newblock Representation engineering: A top-down approach to ai transparency.
\newblock \emph{arXiv preprint arXiv:2310.01405}.

\end{thebibliography}
